\documentclass[preprint,12pt]{elsarticle}

\usepackage{amssymb}
\usepackage{amsmath}
\usepackage{lineno}

\usepackage{hyperref}
\usepackage{algorithmicx}
\usepackage{algorithm}
\usepackage{xcolor}
\usepackage{amsmath}
\usepackage{subfigure}
\usepackage{float}
\usepackage{algpseudocode}
\usepackage{makecell}
\usepackage{comment}
\usepackage{multirow}

\usepackage{graphicx} 
\usepackage{color}
\usepackage{tikz}
\usetikzlibrary{arrows.meta}
\usepackage{amsfonts}

\everymath{\color{blue!80!black}}
\everydisplay{\color{blue!80!black}}
\newcommand{\betaexp}{\psi}
\newcommand{\pcn}{\textsf{pCN}}

\definecolor{Goldenrod}{rgb}{1.0, 0.84, 0.0}
\definecolor{Red}{rgb}{0.77, 0.01, 0.2}
\definecolor{Lavender}{rgb}{0.78, 0.08, 0.52}
\definecolor{MidnightBlue}{rgb}{0.1, 0.1, 0.44}
\definecolor{mrborange}{RGB}{252,187,106}
\definecolor{mrbblue}{RGB}{40,6,76}
\definecolor{mediumchampagne}{rgb}{0.95, 0.9, 0.67}
\definecolor{airforceblue}{rgb}{0.36, 0.54, 0.66}

\newcommand{\Yref}{\Y^{\mathsf{ref}}}
\newcommand{\mynorm}[1]{\parallel #1 \parallel}
\newcommand{\errY}{\mathsf{RE}_{\Y}}
\newcommand{\errYn}[1]{\errY({#1})}
\newcommand{\errpost}{\mathsf{DRE}}
\newcommand{\errpostn}[1]{\errpost({#1})}
\newcommand{\vd}{\vec{v}_{_{{\mathsf{D}}}}}
\newcommand{\pres}{\mathsf{p}}

\newcommand{\Domb}{\Gamma}
\newcommand{\Dom}{\mathcal{D}}

\newcommand{\dkl}{\mathcal{D}_{_{\!\mathcal{KL}}}}
\newcommand{\dklf}[2]{\dkl\left(#1 || #2  \right)}

\newcommand{\vari}[1]{\sigma_{#1}^{2}}
\newcommand{\desv}[1]{{\sigma_{#1}}}
\newcommand{\me}[1]{\mu_{#1}}

\newcommand{\normalf}[2]{\normal\left(#1,#2\right)}

\newcommand{\Nb}{\N_{b}}
\newcommand{\Nz}{\N_{z}}

\newcommand{\ls}{z}
\newcommand{\decoder}{\mathsf{D}}
\newcommand{\encoder}{\mathsf{E}}
\newcommand{\eparam}{\epsilon}
\newcommand{\dparam}{\delta}
\newcommand{\lmse}{\mathcal{L}_{_{\mathsf{rec}}}}
\newcommand{\MSE}{\textsf{MSE}}
\newcommand{\lreg}{\mathcal{L}_{_{\mathsf{reg}}}}
\newcommand{\ltot}{\mathcal{L}_{_{\mathsf{tot}}}}

\newcommand{\prior}{\textit{prior}}
\newcommand{\set}[3]{\left\{ #1 \right\}_{#2}^{#3}}
\newcommand{\dataset}[1]{\mathbb{#1}}
\newcommand{\datael}[1]{\mathsf{#1}}
\newcommand{\dataeln}[2]{\datael{#1}^{(#2)}}

\newcommand{\N}{\mathsf{N}}
\newcommand{\Nx}{\N_{x}}
\newcommand{\iid}{\textit{i.i.d.}}
\newcommand{\fp}[1]{p(#1)}

\newcommand{\fP}[2]{p_{#1}(#2)}
\newcommand{\fQ}[2]{q_{#1}(#2)}

\newcommand{\Np}{\mathsf{Np}}
\newcommand{\Nd}{\mathsf{Nd}}

\newcommand{\identity}{I}
\newcommand{\identityn}[1]{\identity_{{#1}}}

\newcommand{\AR}{\mathsf{AR}}
\newcommand{\eAR}{\widehat{\AR}}

\newcommand{\MT}{\texttt{MT}}

\newcommand{\KS}{\textsf{KS}}

\newcommand{\psrf}{\widehat{\mathsf{R}}}

\newcommand{\MPSRF}{\textsf{MPSRF}}

\newcommand{\covC}{{\boldsymbol{\mbox{$\mathcal{C}$}}}}

\newcommand{\covs}{\boldsymbol{\varSigma}}

\newcommand{\pertub}{\boldsymbol{\varepsilon}}

\newcommand{\VAE}{\textsf{VAE}}
\newcommand{\vaen}[1]{\VAE$_{[\clen = #1]}$}
\newcommand{\vaei}{\vaen{10-30}}
\newcommand{\vaeii}{\vaen{15-35}}
\newcommand{\vaeiii}{\vaen{10-35}}

\newcommand{\vet}[1]{\vec{#1}}
\renewcommand{\dim}{\mathsf{d}}
\newcommand{\minimo}[2]{\min\!\left\{#1,#2 \right\}}

\newcommand{\accept}[2]{\alpha(#1, #2)}

\newcommand{\apriori}{\textit{a priori}}
\newcommand{\maxit}{\mathsf{MaxIter}}

\newcommand{\precistd}{\sigma}
\newcommand{\preci}{\precistd^{2}}

\newcommand{\cprob}[2]{\prob{#1 | #2}}
\newcommand{\prob}[1]{\mathsf{P}\!\left(#1\right)}
\newcommand{\prop}{\mathsf{q}}
\newcommand{\propf}[1]{\prop(#1)}

\newcommand{\post}[1]{\pi(#1)}

\newcommand{\statec}{{\theta}}

\newcommand{\state}{\boldsymbol{\statec}}
\newcommand{\jsize}{\gamma}

\newcommand{\staten}[1]{\state^{{#1}}}
\newcommand{\data}{\pres}
\newcommand{\datap}{\tilde{\data}}
\newcommand{\datan}[1]{\data^{#1}}
\newcommand{\datapn}[1]{\datap^{#1}}

\newcommand{\train}[1]{$\mathbb{D}_{[\clen = #1]}$}

\newcommand{\simul}{\mathsf{sim}}
\newcommand{\rf}{\mathsf{ref}}

\newcommand{\energy}[1]{\mathsf{E}({#1})}

\newcommand{\perm}{\kappa}
\newcommand{\permt}{\boldsymbol{\perm}}

\newcommand{\mm}{\mathrm{m}}

\newcommand{\vtheta}{\boldsymbol{\theta}}

\newcommand{\kl}{\textsf{KL}}
\newcommand{\kle}{\textsf{KLE}}
\newcommand{\klen}[1]{\kle$_{[\clen = #1]}$}
\newcommand{\KL}{{Karhunen-Lo\`eve}\ }
\newcommand{\kldiv}{$\mathcal{KL}$\ }
\newcommand{\Y}{\mathsf{Y}}

\newcommand{\ale}{\omega}
\newcommand{\Yxw}{\Y(\vx,\ale)}
\newcommand{\ava}{\lambda}
\newcommand{\avai}{\ava_{i}}
\newcommand{\avaj}{\ava_{j}}
\newcommand{\ave}{\boldsymbol{\Phi}}
\newcommand{\avei}{\ave_{i}}

\renewcommand{\exp}{\mbox{exp}}
\newcommand{\normal}{N}
\newcommand{\clen}{\ell} 
\newcommand{\D}{\mathcal{D}}
\newcommand{\mcmc}{\textsf{McMC}} 
\newcommand{\apri}{{\em a priori}}
\newcommand{\apost}{{\em a posteriori}}

\newcommand{\events}{\mathcal{A}} 
\newcommand{\probab}{\mathcal{P}} 

\newcommand{\nruns}{\mathsf{NT}} 
\newcommand{\na}{\mathsf{NA}}

\newcommand{\poro}{\phi}  
\newcommand{\fx}{\!\left(\vx\right)}
\newcommand{\fxw}{\!\left(\vx,\ale\right)}

\newcommand{\vx}{\vetor{x}}
\newcommand{\vy}{\vetor{y}}

\newcommand{\visc}{\mu}   

\newcommand{\eq}[1]{\Eq~(\ref{#1})}

\newcommand{\Eq}{Eq.}

\newcommand{\Fig}{Figure}
\newcommand{\Figs}{Figures}
\newcommand{\fig}[1]{\Fig~\ref{#1}}
\newcommand{\figsto}[2]{\Figs~\ref{#1} to \ref{#2}}

\newcommand{\algo}[1]{Algorithm~(\ref{#1})}

\newcommand{\Tab}{Table}

\newcommand{\tab}[1]{\Tab~\ref{#1}}

\newcommand{\Div}[1]{\nabla_{\! {#1}}\cdot}
\newcommand{\grad}{\nabla}

\newcommand{\vetor}[1]{\vet{#1}}
\newcommand{\covv}[1]{\covC_{#1}(\vetor{x}, \vetor{y})}
\newcommand{\covn}[1]{\covC_{#1}}

\newcommand{\real}{\mathbb{R}}

\newcommand{\med}[1]{\langle {#1} \rangle}

\usepackage{booktabs}

\journal{Mathematics and Computers in Simulation}

\begin{document}

\begin{frontmatter}


\title{Variational Autoencoder for Generating Broader-Spectrum \textit{prior} Proposals in Markov chain Monte Carlo Methods}
    
\author[inst1]{Marcio Borges\corref{cor1}}
\cortext[cor1]{Corresponding author.}
\author[inst2]{Felipe Pereira}
\author[inst1]{Michel Tosin}

\affiliation[inst1]{organization={National Laboratory for Scientific Computing},
            addressline={Av. Getulio Vargas, 333, Quitandinha}, 
            city={Petropolis},
            postcode={25651-075}, 
            state={RJ},
            country={Brazil}}

\affiliation[inst2]{organization={Department of Mathematical Sciences, The University of Texas at Dallas},
            addressline={800W Campbell Rd.}, 
            city={Richardson},
            postcode={75080}, 
            state={TX},
            country={USA}}

\begin{abstract}
This study uses a Variational Autoencoder method to enhance the efficiency and applicability of Markov Chain Monte Carlo (\mcmc) methods by generating broader-spectrum prior proposals. Traditional approaches, such as the Karhunen-Loève Expansion (\kle), require previous knowledge of the covariance function, often unavailable in practical applications. The \VAE\ framework enables a data-driven approach to flexibly capture a broader range of correlation structures in Bayesian inverse problems, particularly subsurface flow modeling. The methodology is tested on a synthetic groundwater flow inversion problem, where pressure data is used to estimate permeability fields. Numerical experiments demonstrate that the \VAE-based parameterization achieves comparable accuracy to \kle\ when the correlation length is known and outperforms \kle\ when the assumed correlation length deviates from the true value. Moreover, the \VAE\ approach significantly reduces stochastic dimensionality, improving computational efficiency. The results suggest that leveraging deep generative models in \mcmc\ methods can lead to more adaptable and efficient Bayesian inference in high-dimensional problems.
\end{abstract}

\begin{highlights}
\item Research highlight 1: While the use of generative models in this context is not new, its benefit in relaxing the choice of the covariance function remains unexplored. Therefore, this work's main contribution is to show the advantages of using deep generative models like \VAE\ to provide more flexible and versatile prior distributions. This allows us to relax the assumption that the covariance function of the fields is known \apriori\ (in practical applications, this is not the case). Typically, the few measurements available are insufficient to obtain this information accurately.

\item Research highlight 2: The advantage mentioned in the previous highlight is achieved without reducing the efficiency of the Metropolis algorithm.
\end{highlights}


\end{frontmatter}



\section{Introduction} \label{intro}

The scarcity of information on the heterogeneous hydraulic properties of geological formations prevents a deterministic description of them. Alternatively, a stochastic approach should be adopted based on the few and sparse data available \citep{glimm93}.
To reduce the uncertainties inherent in the stochastic description of the properties of interest, Bayesian methods have introduced dynamic flow data into the models, providing valuable information on the fluids' behavior. This procedure gives rise to a stochastic inverse problem in which the sampling process of the posterior distribution can be performed using the Markov chain Monte Carlo method (\mcmc) \citep{glimm99, casella05, efendiev06, Akbarabadi2015, ALI2024112609}.

The Metropolis algorithm \citep{metropolis1953} and its variants are an important class of \mcmc\ algorithms widely used in Bayesian analysis due to their simplicity and general applicability. \mcmc\ methods are considered the gold standard technique for Bayesian inference \citep{Nemeth2021}. However, in problems with high stochastic dimensions, their convergence can become very slow, possibly making their practical use unfeasible. Below, we describe some measures that can be taken to overcome this drawback.

In porous media problems, forward models are governed by complex partial differential equations (PDEs) whose fine-mesh solutions must be approximated at each iteration of the Metropolis algorithm. Therefore, most of the time is spent simulating the flow problems. Efficient flow simulators are essential tools for mitigating the computational burden. Also, in this sense, multistage (or preconditioned) \mcmc\ methods, which use coarse-mesh models (upscaled) to preselect proposals, can be used to alleviate the computational burden \citep{christenfox05,efendiev2006}.

Stochastic dimension reduction methods are another widely used approach to accelerate the convergence of algorithms. Among them, we can mention the \KL\ (\kl) expansion \citep{karhunen46,loeve55}; Discrete Cosine Transform (DTC) \citep{aleardi_combining_2020, vinciguerra_discrete_2022}; Variational Autoencoders (\VAE) \citep{kingma2014, Xu_2024}; and multi-scale sampling \citep{ALI2024112609}.

In the Metropolis algorithm (and its variants), the selection or adjustment of the proposal distribution is not a trivial task and can become the bottleneck of the entire work \citep{haario2005}. The jump size and the shape of the proposal distribution are crucial for the convergence of the chains and, therefore, for ensuring computational efficiency. In this sense, several methodologies have been proposed to accelerate the convergence of algorithms. Among them are: Adaptive Proposal \citep{haario99}; Adaptive Metropolis \citep{haario2001}; Delayed Rejection \citep{Tierney1994, TierneyMira1999}; Delayed Rejection Adaptive Metropolis algorithm \citep{Haario2006}; Single Component Adaptive Metropolis \citep{haario2005}; Differential Evolution Markov Chain \citep{vrugt2003, terbraak2006}; Differential Evolution Adaptive Metropolis \citep{vrugt2008,vrugt2016}; Domain-decomposed \VAE-\mcmc\ \citep{Xu_2024}, among others.

The prior distribution should reflect our best knowledge about the parameters based on the scarce information available. In Bayesian inverse porous media problems, using \kle\ traditionally requires \apriori\ knowledge of the correlation length in the covariance function. However, this information is often unavailable in most practical applications, and a single covariance model is assumed in an \textit{ad hoc} manner. In contrast, recent data-driven methods, like \VAE\, do not rely on such strict assumptions. Instead, they can use the prior knowledge derived from a diverse training dataset \citep{Xia2022, Xu_2024}. Within the \VAE\ framework, the Metropolis algorithm selects the best model based on the dynamic data incorporated into the likelihood function. Therefore, the central objective of this work is to compare the posterior distributions of the parameter of interest (permeability, or equivalently, the associated Gaussian field) obtained using the Metropolis algorithm with different forms of prior distribution parameterization. To this end, we propose to exploit the \VAE\ method trained with fields with various correlation lengths to generate a broader spectrum prior distribution. The experiments demonstrated that the \VAE\ method yields results comparable to the \kle\ method when a known correlation length is used. Furthermore, it outperforms the \kle\ approach when the correlation lengths applied differ from the true one.

This work is structured as follows: Section~\ref{sec:sthoc} defines the inverse problem being addressed. Section~\ref{sec:perm} presents the parameterization of permeability. Section~\ref{sec:mcmc} describes the Markov chain Monte Carlo (\mcmc) method employed for sampling the posterior distribution. Section~\ref{sec:results} presents the numerical results of the experiments. Finally, Section~\ref{sec:conclusions} provides the conclusions drawn from this work.

\section{Stochastic flow problem}\label{sec:sthoc}

In this section, we present the stochastic steady-state groundwater problem that demonstrates the effectiveness of the proposed methodology. Since the permeability field is unknown, we will use sensor pressure measurements to perform a Bayesian stochastic inversion process and sample its posterior distribution. First, we will outline the equations governing single-phase flow in heterogeneous porous media and then describe the model used to determine permeability.

\subsection{Flow model}\label{sec:model}

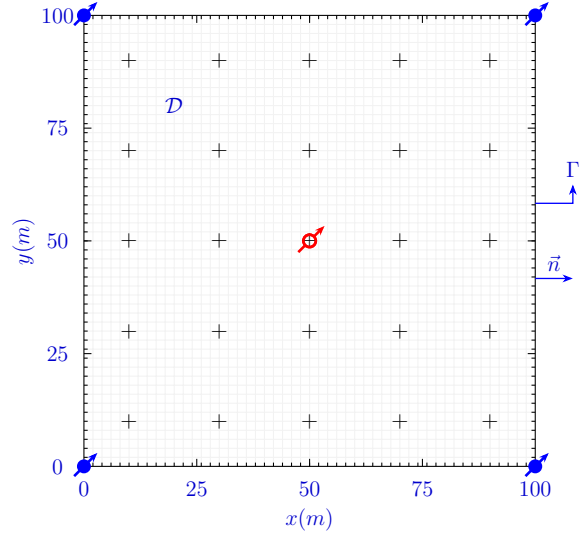
\begin{figure}[H]
    \centering
\begin{tikzpicture}[>=latex,scale=1]

\pgfmathsetmacro{\nodescale}{0.7}
\pgfmathsetmacro{\crad}{0.075}
\pgfmathsetmacro{\atop}{0.18}
\pgfmathsetmacro{\abot}{-0.13}

\draw[step=0.12,thin,help lines,draw=gray!10] (-3,-3) grid (3,3);

\draw[draw=red,line width=1,-{Stealth[scale=0.5]}] (-0.15,-0.15) -- (0.2,0.2);
\draw[draw=red,fill=white,line width=1] (0,0) circle (1.1*\crad);

\foreach \x in {1,2,...,49}{
    \draw[draw=black] (-3+0.12*\x,-3)--(-3+0.12*\x,-2.95);
    \draw[draw=black] (2.95,-3+0.12*\x)--(3,-3+0.12*\x);
    \draw[draw=black] (-3+0.12*\x,2.95)--(-3+0.12*\x,3);
    \draw[draw=black] (-3,-3+0.12*\x)--(-2.95,-3+0.12*\x);
}
\foreach \x in {1,2,3}{
    \draw[draw=black] (-3+1.5*\x,-3)--(-3+1.5*\x,-2.9);
    \draw[draw=black] (2.9,-3+1.5*\x)--(3,-3+1.5*\x);
    \draw[draw=black] (-3+1.5*\x,2.9)--(-3+1.5*\x,3);
    \draw[draw=black] (-3,-3+1.5*\x)--(-2.9,-3+1.5*\x);
}
\foreach \x in {1,2,3,4,5}{
    \node[scale=\nodescale] at (-3.6+1.2*\x,-2.4){+};
    \node[scale=\nodescale] at (-3.6+1.2*\x,-1.2){+};
    \node[scale=\nodescale] at (-3.6+1.2*\x,0){+};
    \node[scale=\nodescale] at (-3.6+1.2*\x,1.2){+};
    \node[scale=\nodescale] at (-3.6+1.2*\x,2.4){+};
    \node[scale=\nodescale] at (-3+1.5*\x-1.5,-3.3){\pgfmathparse{(\x-1)*25}\pgfmathprintnumber[fixed,precision=1]{\pgfmathresult}};
    \node[scale=\nodescale] at (-3.35,-3+1.5*\x-1.5){\pgfmathparse{(\x-1)*25}\pgfmathprintnumber[fixed,precision=1]{\pgfmathresult}};
}

\draw (-3,-3) rectangle (3,3);

\draw[draw=red,line width=1] (0,0) circle (1.1*\crad);

\draw[draw=blue,line width=1,-{Stealth[scale=0.5]}] (\abot-3,\abot-3) -- (\atop-3,\atop-3);
\draw[draw=blue,fill=blue,line width=1] (-3,-3) circle (\crad);
\draw[draw=blue,line width=1,-{Stealth[scale=0.5]}] (\abot-3,\abot+3) -- (\atop-3,\atop+3);
\draw[draw=blue,fill=blue,line width=1] (-3,3) circle (\crad);
\draw[draw=blue,line width=1,-{Stealth[scale=0.5]}] (\abot+3,\abot+3) -- (\atop+3,\atop+3);
\draw[draw=blue,fill=blue,line width=1] (3,3) circle (\crad);
\draw[draw=blue,line width=1,-{Stealth[scale=0.5]}] (\abot+3,\abot-3) -- (\atop+3,\atop-3);
\draw[draw=blue,fill=blue,line width=1] (3,-3) circle (\crad);

\draw[draw=blue,line width=0.5,-{Stealth[scale=0.75]}] (3,-0.5) -- (3.5,-0.5);
\node[scale=\nodescale,above] at (3.25,-0.5){$\vetor{n}$};

\draw[draw=blue,line width=0.5,-{Stealth[scale=0.75]}] (3,0.5) -- (3.5,0.5) -- (3.5,0.75);
\node[scale=\nodescale,above] at (3.5,0.75){$\Domb$};
\node[scale=\nodescale] at (-1.8,1.8){$\Dom$};
\node[scale=\nodescale] at (0,-3.7){$x(m)$};
\node[scale=\nodescale,rotate=90] at (-3.8,0){$y(m)$};
\draw[draw=white] (3.8,-2) -- (3.8,2);

\end{tikzpicture}

    \caption{Simulation domain. The injection well is indicated in red at the center of the domain, while the production wells are marked in blue. Black crosses represent pressure sensors.}
    \label{fig:domain}
\end{figure}

Let $\Dom \subset \real^{2}$ be the domain, with boundary $\Domb$ and unit outward normal $\vetor{n}$, occupied by a heterogeneous and rigid porous media saturated by water. Assuming a homogeneous porosity ($\poro = 0.2$) and denoting $\vd$ the Darcy velocity, the single-phase flow of an incompressible fluid, in the absence of gravity, is described by the equations:

\vspace{6pt}
\begin{center}
\begin{equation}
  \Div{}\vd\fx = \vetor{q}\fx, \quad \mbox{\textcolor{black}{and}} \quad \vd\fx = -\dfrac{\permt\fx}{\visc} \grad \pres\fx,
  \label{eq:darcy}
\end{equation}
\end{center}
\vspace{6pt}

\noindent where $\vetor{q}$ is the source term, $\pres$ and $\visc$ are the fluid's pressure and viscosity, respectively. Under the assumption of isotropy, the permeability tensor $\permt$ is treated as a scalar ($\perm$). The square two-dimensional domain ($\Dom = [0,100]\times [0,100]\mathsf{m}^{2}$) is discretized in a regular mesh of $50 \times 50$ elements (\fig{fig:domain}). The permeability is piecewise constant in each element.

We consider a five-spot well arrangement with homogeneous Neumann boundary conditions. A Peaceman-type model represents these wells (\fig{fig:domain}). A flow rate ($\vetor{q}=100\ \mathsf{m}^3 / \mathsf{day}$) is imposed on the injection well, and a constant bottom hole pressure ($\pres_{bh} = 1.01325\times 10^{5}\ \mathsf{Pa}$) on the production wells. The water viscosity is equal $10^{-3}\mathsf{Pa}\cdot\mathsf{s}$. A two-point flux approximation (TPFA) scheme approximates the solution of the mathematical problem formed by \eq{eq:darcy} using the MATLAB/OCTAVE Reservoir Simulation Toolbox (MRST) simulator from Sintef \citep{matlabbook}.

\subsection{Permeability modeling}

In this work, the permeability field, $\perm(\vx,\ale)$, is treated as a random space function with statistics inferred from geostatistical models.
Here $\vx =\left( x_{1},x_{2}  \right)^{\! ^{\mathsf{T}}}\in \mathbb{R}^{2}$ and $\ale$ is a random element in the probability space.
In line with \citet{dagan89} and \citet{gelhar93}, the permeability field is modeled as a log-normally distributed function
\vspace{6pt}
\begin{center}
\begin{equation}
    \perm\fxw = \betaexp \, \exp\Big[\rho\Y\fxw\Big],
    \label{field}
\end{equation}
\end{center}
\vspace{6pt}
\noindent where $\betaexp,\rho\in\mathbb{R}^{+}$ and $\Y\fxw \sim \normalf{\me{\Y}}{\covn{\Y}}$ is a Gaussian random field characterized by its mean $\me{\Y} = \med{\Y}$ and two-point covariance function $\covv{\Y}$. Here, we consider a squared exponential covariance function 

\begin{equation}
    \covv{\Y} = \sigma^{2}_{\Y}\ \exp\left( -\dfrac{|x_{1} - y_{1}|^{2}}{2\ell^2_{1}} -\dfrac{|x_{2} - y_{2}|^{2}}{2\ell^2_{2}} \right),
    \label{eq:cov}
\end{equation}

\noindent with $\sigma^{2}_{\Y}$ denoting the variance and $\ell_{i}>0 \ (i=1,2)$ the correlation lengths. In our studies, we set $\betaexp = 9.87\times 10^{-14}\ \mathsf{m}^2$, $\rho = 1.0$, and for simplicity, $\ell_{1} = \ell_{2} = \ell$.

\section{Permeability parameterization}\label{sec:perm}

For the stochastic inversion process used in this work, we need to simulate the correlated fields $\Y\fxw$ from \eq{field} (the \apri\ distribution) stochastically. Several methods have been developed to achieve this, including the Spectral Representation Method, Sequential Gaussian Simulation, LU Decomposition Algorithm, Turning Band Algorithm, and Simulated Annealing (see \citet{deutsch1992gslib} and references therein). This work focuses on two parameterization methods for addressing high-dimensionality problems: the \KL\ expansion and the Variational Autoencoder. It is worth noting that both allow for reducing the stochastic dimension, a fundamental characteristic in applications with \mcmc\ methods. Next, we present the formulation of these methods.

\subsection{\KL expansion (\kle)}\label{sec:KLE}

The Gaussian field $\Y$ can be represented as a series expansion involving a complete set of deterministic functions with correspondent random coefficients using the \KL\ (\kl) expansion proposed independently by \citet{karhunen46} and \citet{loeve55}. 
It is based on the eigen-decomposition of the covariance function. Depending on how fast the eigenvalues decay, one may be able to retain only a small number of terms in a truncated expansion. As a result, this approach may narrow the search to a smaller parameter space. In uncertainty quantification methods for porous media flows, the \kle\ has been widely used to represent the permeability field \citep{efendiev05,efendiev2006,das10,mondal10,ginting11,ginting12}.
Another advantage of \kle\ lies in the fact that it provides orthogonal deterministic basis functions and uncorrelated random coefficients, allowing for the optimal encapsulation of the information contained in the random process into a set of discrete uncorrelated random variables \citep{GhanemSpanos}. This remarkable feature can be used to simplify the Metropolis-Hastings \mcmc\ Algorithm in the sense that the search may be performed in the space of discrete uncorrelated random variables ($\vtheta$), no longer in the space of permeabilities which have a more complex statistical structure.

Now, we recall the essential elements of the \kle. Consider $\Y(\vx,\ale)$ defined on a probability space $(\Omega,\events,\probab)$ composed by the sample space, the ensemble of events, and a probability measure, respectively, and indexed on a bounded domain $\D \subset \mathbb{R}^{2}$ (defined in Section~\ref{sec:sthoc}).
The process $\Y$ can be expressed as 
\begin{equation}
  \Yxw = \med{\Y(\vx)} +\sum_{i=1}^{\infty}\sqrt{\avai} \avei(\vx)
\theta_{i}(\ale),
\label{kl}
\end{equation}
\noindent where $\theta_{i}(\ale)$ is a set of independent random variables; $\avai$ and $\avei$ are the eigenvalues and the square-integrable orthogonal eigenfunctions of the covariance function $\covv{\Y}$, respectively. By definition, $\covv{\Y}$ is bounded, symmetric, and positive definite.

The eigenvalues and eigenfunctions of \eq{kl} are the solution of the homogeneous Fredholm integral equation of the second kind given by
\begin{equation}
  \int_{\D} \covv{\Y} \ave(\vx) d\vx = \ava\ave(\vy).
\label{fred}
\end{equation}

By arranging the eigenvalues in descending order, the $\mm$ truncated series is given by 
\begin{equation}
  \Y\fxw \approx \med{\Y(\vx)} +\sum_{i=1}^{\mm}\sqrt{\avai} \avei(\vx)
\theta_{i}(\ale).
\label{klM}
\end{equation}

The number of terms in the series is determined by the partial energy carried by a subset of the eigenvalues, as defined by the following expression:

\begin{equation}
 \energy{\mm} = \dfrac{\sum_{i=1}^{\mm}\avai}{\sum_{j=1}^{n\rightarrow\infty}\avaj}.
 \label{eq:energy}
\end{equation}

\subsection{Variational Autoencoder \VAE}\label{sec:VAE}

The variational autoencoder (\VAE) model is a stochastic inference and learning algorithm based on variational Bayes (VB) inference proposed by \citet{kingma2014}. Unlike a standard autoencoder, \VAE\ provides a distribution-based latent data representation. It encodes the input dataset $\dataset{X}$ into a probability distribution (in a latent space, characterized by the variable $\ls$). It reconstructs (decodes) the original input using samples from $\ls$. This generative model enforces a \prior\ on the low-dimensional latent space, which is mapped back into a realistic-looking image. Therefore, the essential characteristic of \VAE\ method, in the context of Monte Carlo methods with Markov chains, is their ability to represent high-dimensional parametric spaces in a low-dimensional latent space \citep{Xia2022, Xu_2024}.

Recent advancements in deep learning have provided new insights into reconstructing porous media \citep{Zhang2022, Zhang2022b, Xu_2024}). \citet{Zhang2022} introduced a reconstruction method that leverages Variational Autoencoders (\VAE) and Fisher information, utilizing high-quality data obtained from CT scanners, thereby achieving good efficiency. On the other hand, \citet{Xu_2024} developed a domain-decomposed variational auto-encoder combined with a Markov chain Monte Carlo approach. This method employs local generative models within smaller subdomains to enhance reconstruction accuracy. \citet{Xia2022} introduced a multiscale Bayesian inference method using a multiscale deep generative model (MDGM). These generative models offer a flexible representation and enable hierarchical parameter generation from coarse to fine scales. They combine the multiscale generative model with \mcmc\ to efficiently obtain posterior parameter samples at different scales.

Along with reducing the stochastic dimension, another advantage of using variational autoencoders is that this method can represent broader prior distributions since it can be trained with various fields with different statistical properties. We will exploit this outstanding feature in this work. According to \citet{Xia2022}, the strong assumptions about the mean and covariance function may cause the \kle\ to fail to accurately reflect the true field’s statistical properties.

\begin{figure}[htbp]
\vspace{3mm}
 \centering
 \scalebox{0.9}{
 \begin{tikzpicture}[>=latex,scale=1]


\definecolor{mygreen}{RGB}{155, 208, 103};
\definecolor{myred}{RGB}{246, 170, 144};
\definecolor{myblue}{RGB}{186, 223, 218};
\definecolor{mygreen2}{rgb}{0,0.5,0};
\definecolor{myblue2}{rgb}{0.39, 0.58, 0.93};
\pgfmathsetmacro{\nodescale}{0.7}

\setlength{\fboxsep}{0pt}
\setlength{\fboxrule}{0.3pt}

\draw[rounded corners=3,line width=0.7,fill=myblue2!50,draw=myblue2!110] (-7.1,-2.2) rectangle (-4.1,2.2);
\node at (-5.9,-2*0.15) {\fbox{\includegraphics[width=2cm]{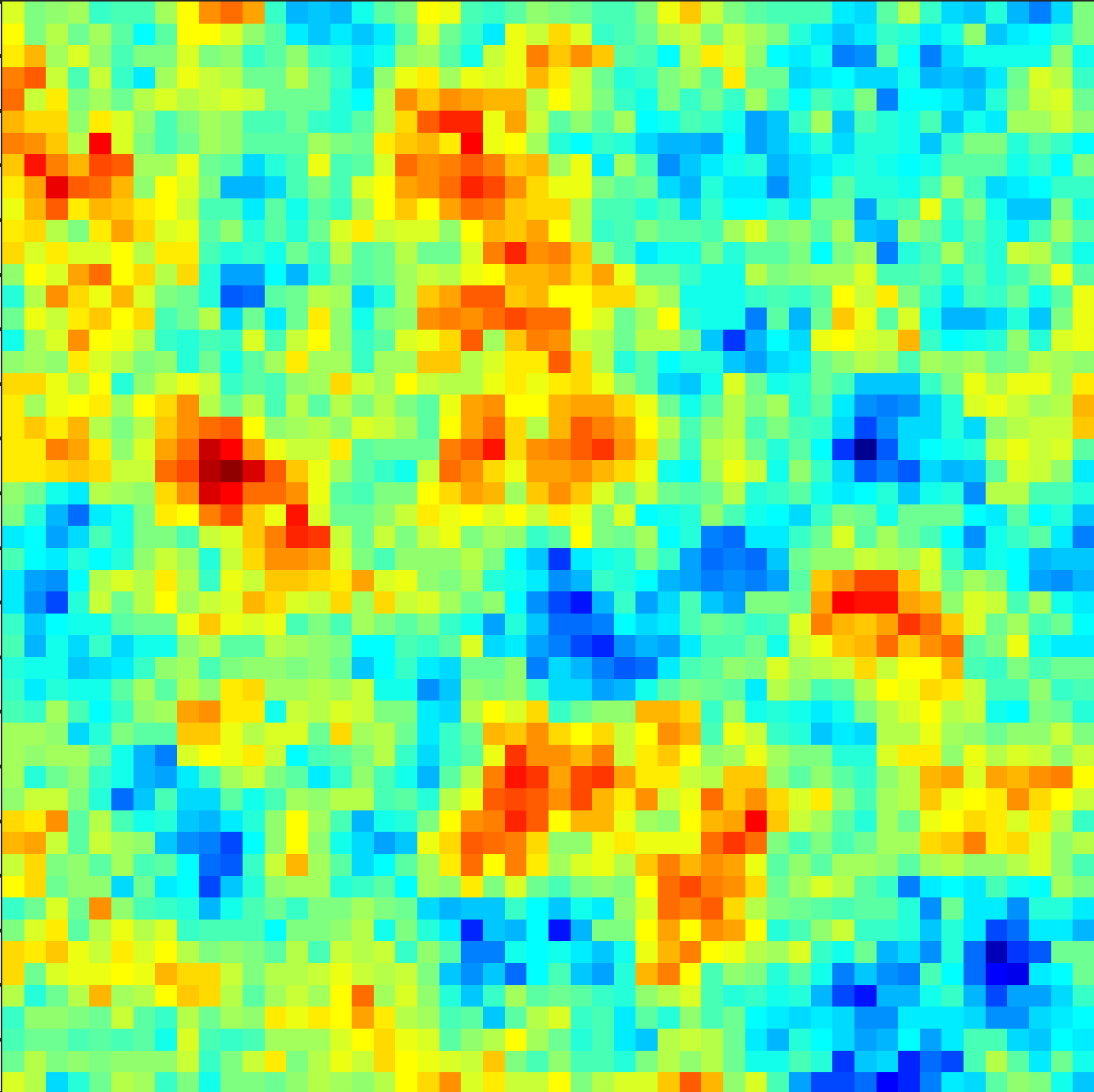}}};
\node at (-5.75,-0.15) {\fbox{\includegraphics[width=2cm]{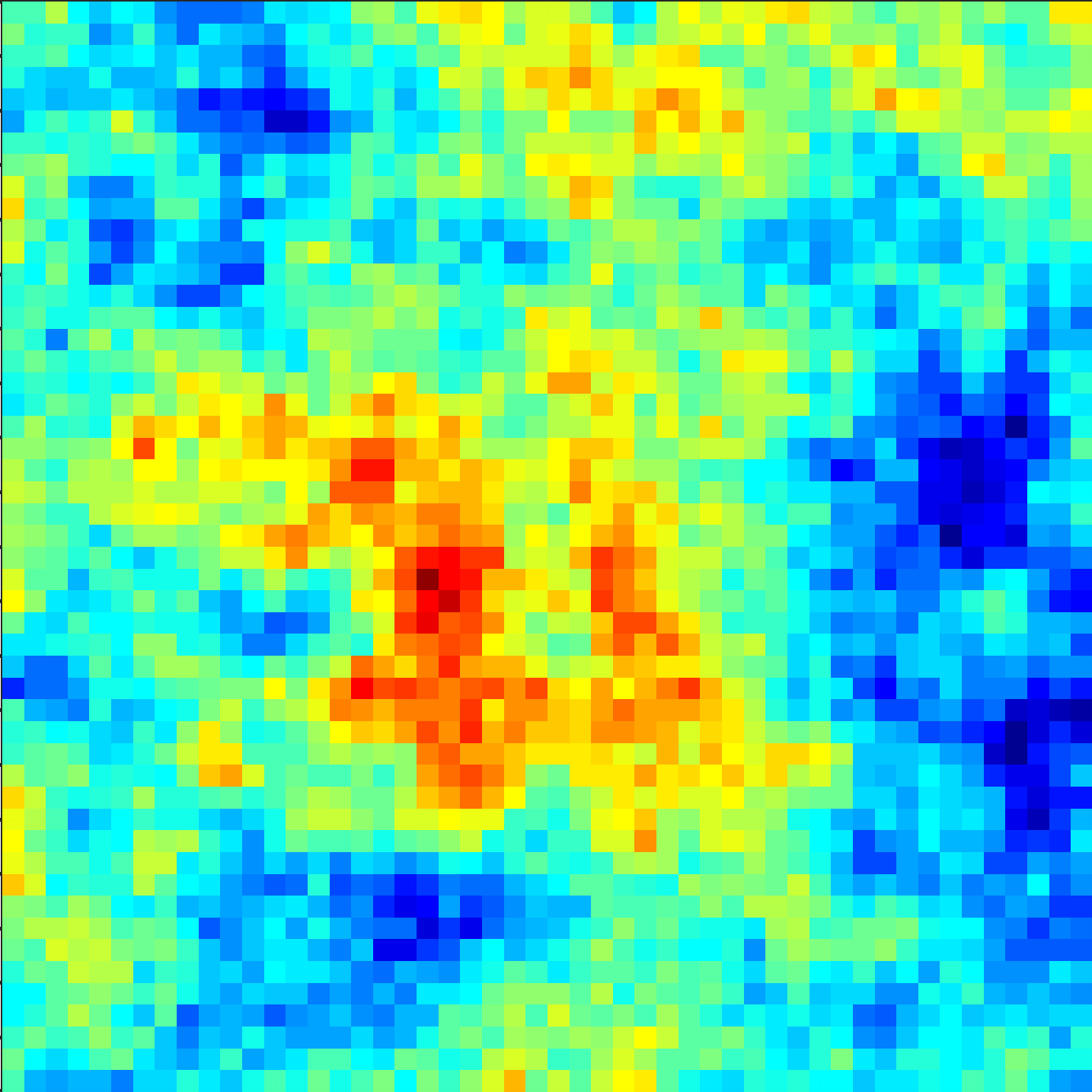}}};
\node at (-5.6,0) {\fbox{\includegraphics[width=2cm]{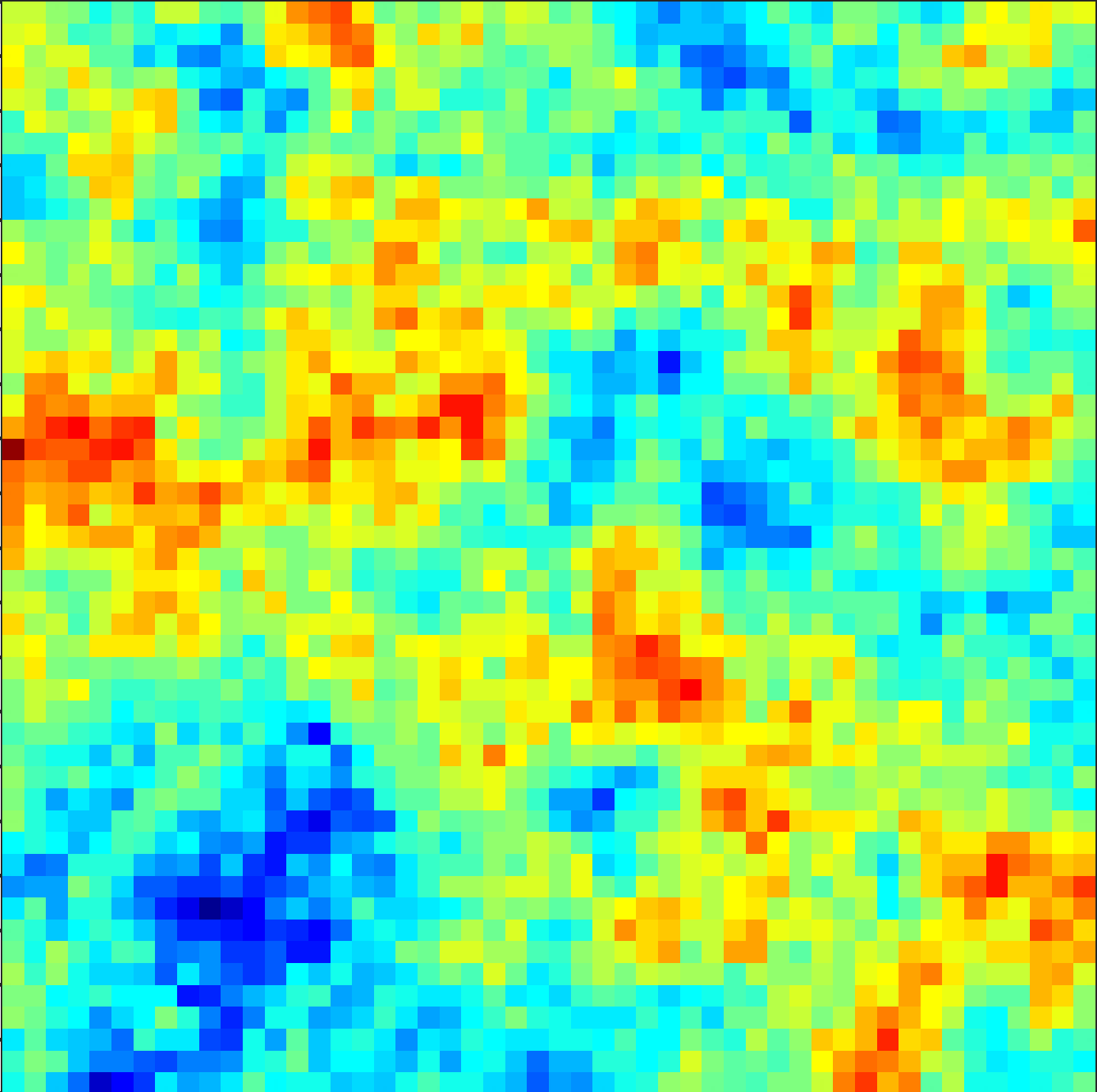}}};
\node at (-5.45,0.15) {\fbox{\includegraphics[width=2cm]{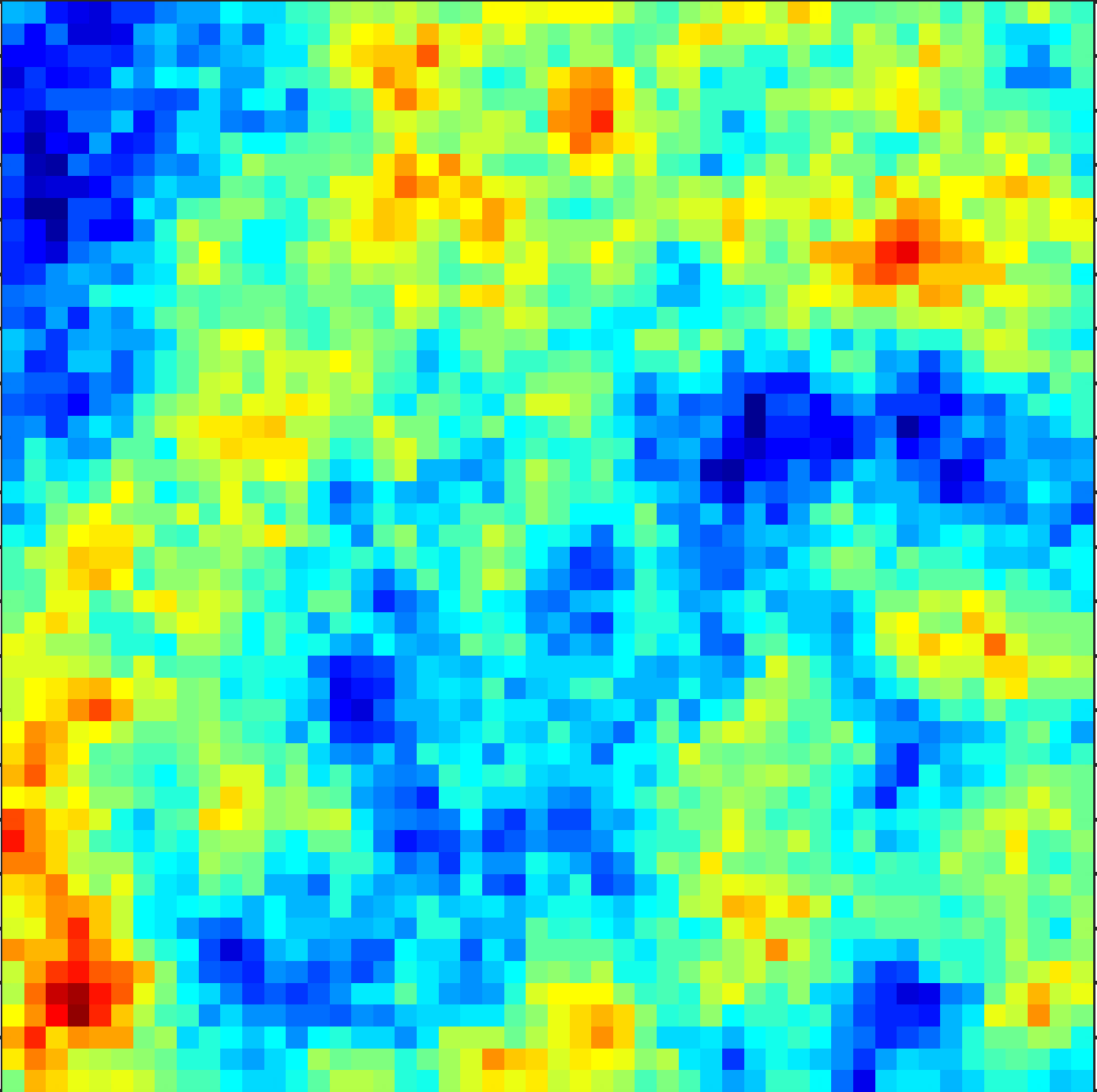}}};
\node at (-5.3,2*0.15) {\fbox{\includegraphics[width=2cm]{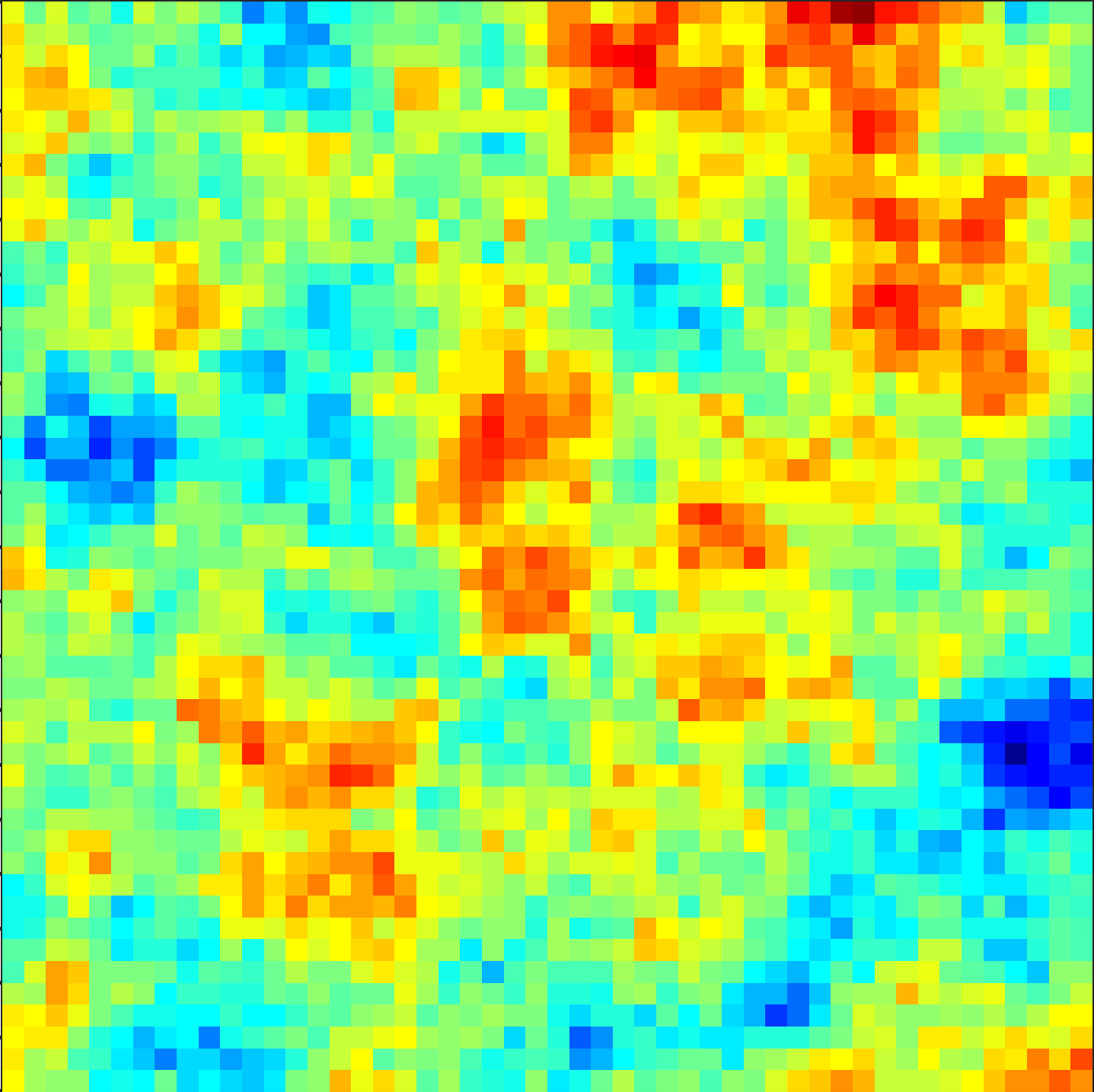}}};
\node[scale=\nodescale] at (-5.6,1.8)[align=center] {Input data};
\node[scale=\nodescale] at (-5.6,-1.8)[align=center] {$\dataset{X}$};
\draw[-{Stealth},line width=0.7] (-4,0) -- (-3.6,0);

\draw[rounded corners=1,line width=0.7,fill=mygreen!50,draw=mygreen] (-3.5,1.4) -- (-3.5,-1.4) 
               -- (-1.6,-0.7)
               -- (-1.6,0.7) -- cycle;
\node[scale=\nodescale] at (-2.55,0)[align=center] {Probabilistic \\[-2.5mm] encoder};
\draw[-{Stealth},line width=0.7] (-1.5,0) -- (-1.2,0) -- (-1.2,1) -- (-0.9,1);
\draw[-{Stealth},line width=0.7] (-1.5,0) -- (-1.2,0) -- (-1.2,-1) -- (-0.9,-1);

\draw[rounded corners=1,line width=0.7,fill=mygreen!50,draw=mygreen] (-0.8,0.6) rectangle (-0.4,1.4);
\draw[rounded corners=1,line width=0.7,fill=myred!40,draw=myred!75] (-0.8,-0.4) rectangle (-0.4,0.4);
\draw[rounded corners=1,line width=0.7,fill=mygreen!50,draw=mygreen] (-0.8,-1.4) rectangle (-0.4,-0.6);
\node[scale=\nodescale] at (-0.6,1) {$\me{\ls}$};
\node[scale=\nodescale] at (-0.6,0) {$\varepsilon$};
\node[scale=\nodescale] at (-0.6,-1) {$\desv{\ls}$};

\draw[-{Stealth},line width=0.7] (-0.3,1) -- (0,1) -- (0,0) -- (0.3,0);
\draw[-{Stealth},line width=0.7] (-0.3,-1) -- (0,-1) -- (0,0) -- (0.3,0);
\draw[-{Stealth},line width=0.7] (-0.3,0) -- (0.3,0);

\draw[rounded corners=1,line width=0.7,fill=myred!50,draw=myred] (0.4,-0.4) rectangle (0.8,0.4);
\node[scale=\nodescale] at (0,2)[align=center] {Sampled latent \\[-2.5mm] variable};
\node[scale=\nodescale] at (0.6,0) {$\ls$};
\node at (0.6,-0.8) {\includegraphics[width=0.7cm]{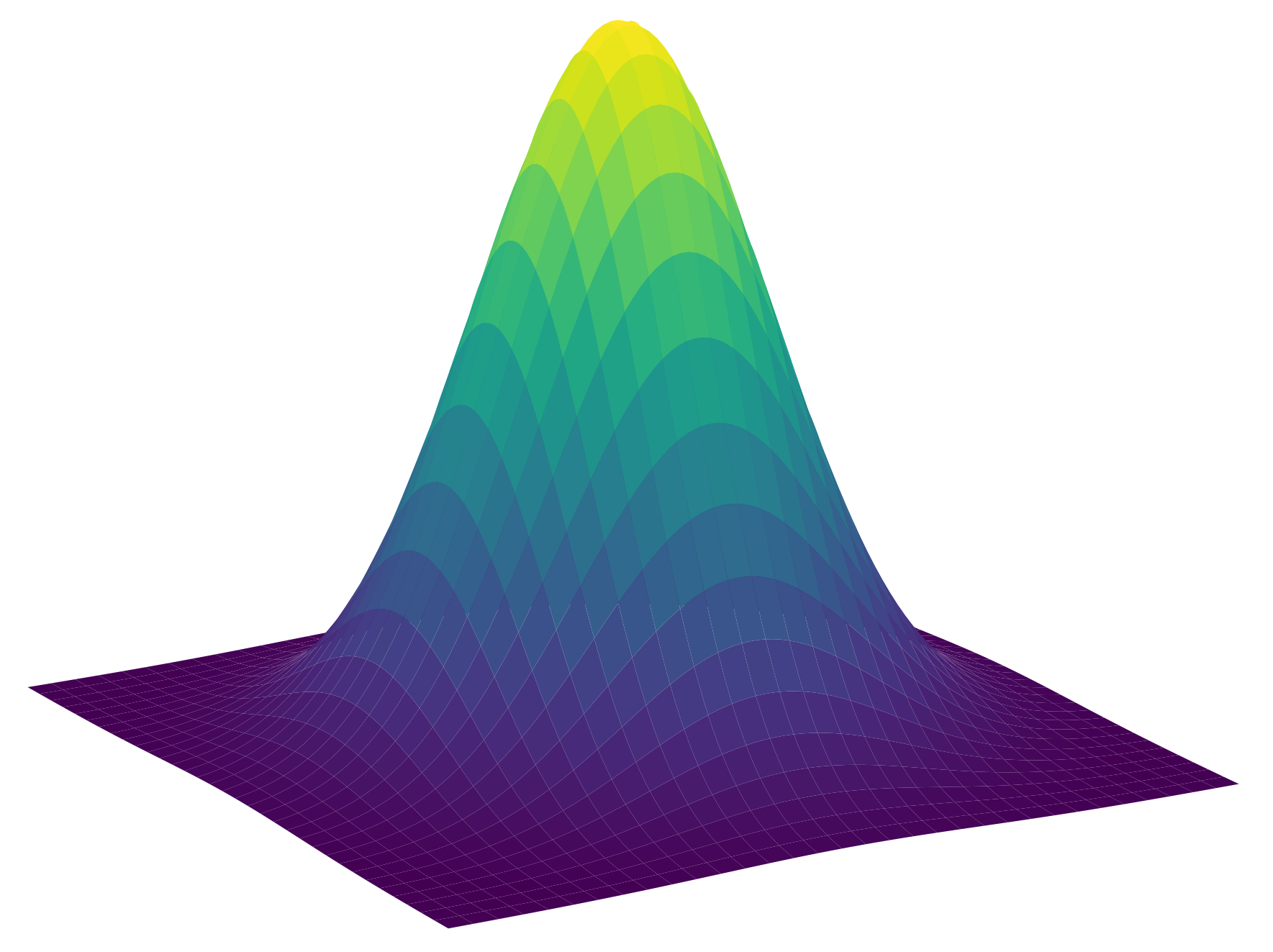}};

\draw[-{Stealth},line width=0.7] (0.9,0) -- (1.5,0);
\draw[rounded corners=1,line width=0.7,fill=myblue!75,draw=myblue!125] (3.5,1.4) -- (3.5,-1.4) 
               -- (1.6,-0.7)
               -- (1.6,0.7) -- cycle;
\node[scale=\nodescale] at (2.55,0)[align=center] {Probabilistic \\[-2.5mm] decoder};

\draw[-{Stealth},line width=0.7] (3.6,0) -- (4,0);
\draw[rounded corners=3,line width=0.7,fill=myblue2!40,draw=myblue2] (4.1,-2.2) rectangle (7.1,2.2);
\node at (5.9,-2*0.15) {\fbox{\includegraphics[width=2cm]{./y4.png}}};
\node at (5.75,-0.15) {\fbox{\includegraphics[width=2cm]{./y3.png}}};
\node at (5.6,0) {\fbox{\includegraphics[width=2cm]{./y2.png}}};
\node at (5.45,0.15) {\fbox{\includegraphics[width=2cm]{./y1.png}}};
\node at (5.3,2*0.15) {\fbox{\includegraphics[width=2cm]{./y0.png}}};
\node[scale=\nodescale] at (5.6,1.8)[align=center] {Reconstructed data};
\node[scale=\nodescale] at (5.6,-1.8)[align=center] {$\dataset{X}'$};

\end{tikzpicture}}
 \label{fig:vae_diag}
 \caption{Diagram representation of the variational autoencoder flow for a Gaussian field reconstruction.}
\end{figure}

Consider the input data set $\dataset{X} = \set{\dataeln{x}{i}}{i=1}{\N}$ ($\dataeln{x}{i}\in\real^{\Nx}$, where $\Nx$ is the number of elements of the field) consisting of $\N$ independent and identically distributed (\iid) samples of the continuous (or discrete) variable drawn from the \prior\ distribution $\fp{\datael{x}}$. The idea is to approximate this distribution by $\fp{\datael{x}|\dparam}$, defined as 

\begin{equation}
    \fp{\datael{x}|\dparam} = \int \fP{\dparam}{\datael{x}|\ls}\fp{\ls} d\ls \:,
\end{equation}

\noindent where $\fp{\ls}$ is the distribution for the latent variable $\ls \in \mathbb{R}^{\Nz}$ ($\Nz < \Nx$) and $\fP{\dparam}{\datael{x}|\ls}$ is a generative model parametrized by $\dparam$. Since this problem is intractable, the solution is to include a variational approximation $\fQ{\eparam}{\ls|\datael{x}}$ that converts the original problem into an optimization one. Further mathematical details are well described in the work of \citet{Xia2022}. The operators $\fQ{\eparam}{\ls|\datael{x}}$ and $\fP{\dparam}{\datael{x}|\ls}$ are called the probabilistic encoder and the probabilistic decoder, respectively. 
If assuming $\fQ{\eparam}{\ls|\datael{x}} = \normalf{\mu_{\ls}(x)}{\sigma^2_{\ls}(x)}$, one 
can apply the reparameterization trick to decomposed $\ls$ as  

\begin{equation}
 \ls = \me{\ls} + \desv{\ls} \odot \varepsilon, 
\end{equation}

\noindent where $\varepsilon \sim \normalf{\vetor{0}}{\identityn{\Nz}}$ and $\odot$ represents the element-wise product \citep{kingma2014}. Now, $\me{\ls}(\datael{x})$ and $\desv{\ls}(\datael{x})$ are calculated by an encoder neural network indicated by 
$\encoder_{\eparam}(x)$ (which also depends of the $\varepsilon$ value to generates $\ls$, but this dependence will 
not be explicitly represented). The decoder neural network, $\decoder_{\dparam}(\ls)$, will return an approximation $x'$ for the field $x$. This process is summarized in the diagram of the \fig{fig:vae_diag}. In addition, this work assumes $\fp{\ls}=\normalf{\vetor{0}}{\identityn{\Nz}}$ and $\fP{\dparam}{\datael{x}|\ls} = \normalf{\decoder_{\dparam}(\ls)}{\identityn{\Nx}}$. 


The loss function of the \VAE\ comprises a reconstruction loss term and a regularization term. 
The first one ensures that the reconstructed image at the output is close (enough) to the input one. 
Here, is given by the mean squared error (\MSE), with the training set organized into batches of size $\Nb$. 
So, for each field, the reconstruction loss is calculated as follows:

\begin{equation}
 \lmse(\eparam, \dparam, \datael{x}) = \left[ \datael{x} - \decoder_{\dparam}\left( \ls \right) \right]^2 .
\end{equation}



\noindent To keep the encoder outputs $\ls$ close to a standard normal distribution and sufficiently diverse, the Kullback–Leibler divergence ($\dkl$, also called relative entropy and I-divergence) is used as a regularization term. The value $\dkl$ is a divergence measure between two distributions \citep{KLD1951,csiszar1975}. In the particular case of interest,

{\small
\begin{equation}
 \lreg(\eparam,x) = \dklf{\fp{\ls}}{{\normalf{0}{1}}} = -\dfrac{1}{2} \displaystyle\sum_{i=1}^{\Nz} \left[1 + \log \left(\vari{\ls_{i}}\right) -  \me{\ls_{i}}^{2}  - \vari{\ls_{i}} \right] \:.
 \label{eq:dkl}
\end{equation}
}

To manage the balance between the two terms, a hyperparameter $\beta$ is commonly included. \citet{higgins2017betavae} documented its effects.
Finally, without loss of generality, the resulting total loss for a batch $\dataset{X}_b$ is then

\begin{equation}
     \ltot\left(\eparam, \dparam, \dataset{X}_b \right) = \dfrac{1}{\Nb} 
                                                \displaystyle \sum_{i=1}^{\Nb} \left[ 
                                                \lmse\left(\eparam, \dparam, \dataeln{x}{i}\right) 
                                                + \beta \lreg\left(\eparam,\dataeln{x}{i}\right) 
                                                \right] \:.
\end{equation}

\section{Markov chain Monte Carlo method (\mcmc)}\label{sec:mcmc}

Markov chain Monte Carlo (\mcmc) algorithms are among the most essential tools for Bayesian data analysis. Although computationally expensive, \mcmc\ methods can handle complex nonlinear problems \citep{xu2020}. Their power lies in their simplicity and the broad applicability of the basic algorithm \citep{haario2005}. The Metropolis algorithm (\MT) was initially introduced by \citet{metropolis1953} for computing properties of substances composed of interacting individual molecules. \citet{hastings70} introduced a generalization to non-symmetric proposals. This algorithm has been widely used in several areas of science.

Let $\post{\cdot}$ be the target distribution (posterior distribution) and $\propf{\staten{t},\state}$ the instrumental proposal distribution. The \MT\ algorithm is given in the \algo{metro}.

{
\begin{algorithm}[H]
  \small
  \caption{Metropolis \mcmc\ Algorithm (\MT) \citep{metropolis1953}}\label{metro}
  \begin{algorithmic}[1]
    \Procedure{Metropolis}{$\maxit$}\Comment{$\maxit$: maximum number of iterations}
      \State \textbf{Initialization:} Generate the initial state $\staten{1}$ from \apriori\ distribution
      \For{$t = 1$ to $\maxit$}
        \State \textbf{Step 1.} At state $\staten{t}$ generate $\state$ from the proposal distribution $\propf{\staten{t},\state}$
        \State \textbf{Step 2.} Take the new state as
        \begin{equation*}
         \staten{t+1} = \left\{
         \begin{array}{ll}
          \state,     & \mbox{ with probability } \accept{\staten{t}}{\state} \\
          \staten{t}, & \mbox{ with probability } 1 - \accept{\staten{t}}{\state}
         \end{array}
         \right. ,
        \end{equation*}
         \hspace{1cm} where 
         \begin{equation}
             \accept{\staten{t}}{\state} =\minimo{1}{\dfrac{\post{\state}}{\post{\staten{t}}}}.
             \label{acceptprob}
         \end{equation}
         \vspace{-0.5cm}
      \EndFor
      \State \Return $\{ \staten{1},\dots,\staten{\maxit} \}$
    \EndProcedure
  \end{algorithmic}
    \label{metroplis_algorithm}
\end{algorithm}
}

\subsection{Likelihood function}

The pressure data (measurements), denoted by $\datan{\rf}$, are combined with the \apriori\ distribution ($\prob{\state}$) through Bayes' theorem to give
\begin{equation}
 \post{\state} = \cprob{\state}{\datan{\rf}} \varpropto \cprob{\datan{\rf}}{\state} \prob{\state},
 \label{eq:post}
\end{equation}

\noindent that is the target (\apost) distribution of the $\dim$-dimensional parameter $\state$.
Before starting the process, the \apriori\ distribution reflects our best knowledge of the parameters.
Assuming that the error between the reference and simulated data ($\datan{\simul}$) has a normal distribution, the likelihood is approximated as

\begin{equation}
    \cprob{\datan{\rf}}{\state} \varpropto \exp\left( - \dfrac{\mynorm{\datan{\simul} - \datan{\rf}}}{\preci} \right).
\end{equation}

\noindent Here, $\preci$ is the overall precision associated with measurement, numerical, and modeling errors.
The numerator of the term within the exponential function is approximated numerically by

\begin{equation}
    \mynorm{\datan{\rf} - \datan{\simul}} = \dfrac{\displaystyle \sum_{i=1}^{\Nd} \left[  \datan{\rf}\!\left(\vx_{i} \right) - \datan{\simul}\!\left(\vx_{i} \right) \right]^{2}}{\displaystyle \sum_{i=1}^{\Nd} \left[  \datan{\rf}\!\left(\vx_{i} \right) \right]^{2}},
    \label{eq:erro}
\end{equation}

\noindent where $\Nd$ is the number of data in space (sensors at \fig{fig:domain}).


The Metropolis algorithm can be used when the \apriori\ knowledge of target distribution is quite limited. However, the selection or ``tuning'' of the proposal distribution may be the bottleneck of the method since a bad choice can lead to a slow convergence rate \citep{gilks1996,haario99}.

In the Metropolis algorithm, a proposed value $\state$ is generated from some pre-established density function $\prop$, which is then accepted with probability $\accept{\cdot}{\cdot}$ following \algo{metro}. The function $\prop$ is typically chosen from some distribution family (e.g., a normal distribution centered at $\staten{t}$). The shape and size of the proposal distribution $\propf{\cdot}$ are crucial for the algorithm's performance and the convergence of the Markov chain \citep{haario99,roberts2001}.

In this work, we use the first-order autoregressive proposal, known as Crank-Nicolson proposal (\pcn) \citep{ALMEIDA20121511,cotter2013, HU2017492, xu2020, Ali2021ConditioningBP} that is a slight variation of the random walk sampler that can lead to significant speed-up to large dimensions and is given by

\begin{equation}
  \state = \left( \sqrt{1 - \jsize^2} \right)\staten{t} + \jsize\pertub,
  \label{eq:autorw}
\end{equation}

\noindent where $\pertub \sim \normalf{\vetor{0}}{\covs}$ is a random perturbation independent of the chain and $\jsize \in(0,1]$ is the tuning parameter associated with the jump. In this case, the probability of acceptance (\eq{acceptprob} at Algorithm~\ref{metroplis_algorithm}) depends only on the likelihood. In the specific case of this work, we consider that initial state $\staten{1}$ (\apriori\ distribution) and perturbation $\pertub$ are both given by a normal distribution, i.e., $\staten{1} \sim \normalf{\vetor{0}}{\identityn{\dim}}$ and $\pertub \sim \normalf{\vetor{0}}{\identityn{\dim}}$, with $\identityn{\dim}$ denoting the $\dim$-dimensional identity matrix. After conducting several numerical experiments, we have chosen a $\jsize$ value equal to $0.1$ for all cases because this value gives the best convergence rate for the problems considered here.

\section{Numerical results} \label{sec:results}

The main goal of this work is to demonstrate that utilizing variational autoencoders to parameterize the prior distribution (specifically, the permeability field in this case) within Markov chain Monte Carlo methods allows us to relax the requirement of fully knowing the covariance function of the fields in advance. By employing neural networks trained on a set of fields with various covariances functions, we can overcome the limitation of proposing a single function, which is commonly done using the Karhunen-Loève expansion. This method enables the \mcmc\ algorithm to select the most suitable fields based on the available data. Furthermore, the \VAE\ method significantly reduces the stochastic dimension, specifically in the present work, from $2,\!500$ to $64$.

To illustrate this approach, we use the Metropolis \algo{metro} to carry out the inversion process for the stochastic flow problem defined in Section~\ref{sec:sthoc}. We aim to sample the posterior distribution of the permeability field using $25$ pressure measurements. We performed six experiments, for which the only difference is in permeability parameterization, as described in the following sections.

\subsection{Experimental setup} \label{sec:setup}

\begin{figure}[htbp]
    \centering
    \subfigure[Gaussian field]{\includegraphics[width=0.32\linewidth]{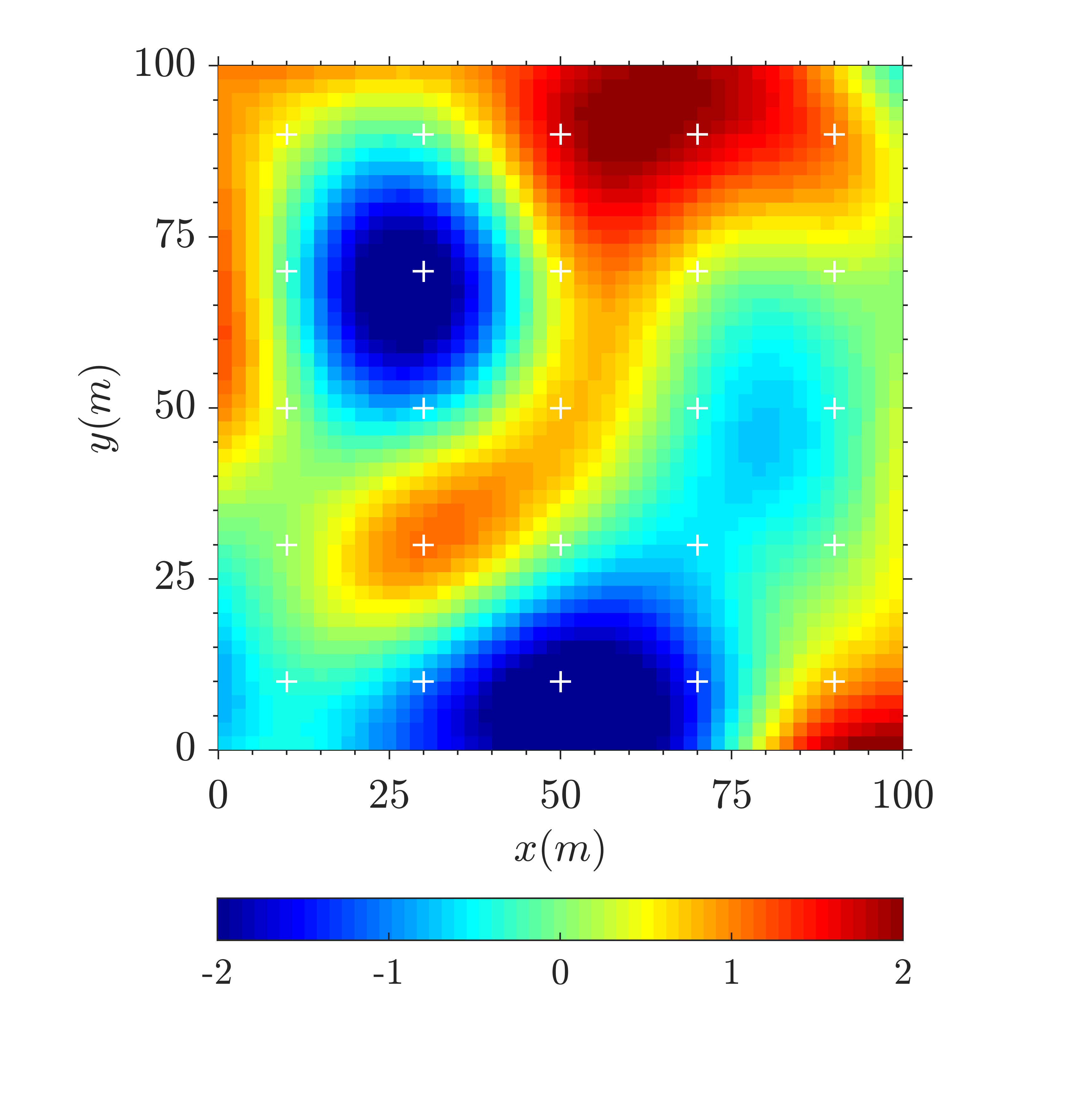}\label{fig:refsa}}
    \subfigure[Permeability field]{\includegraphics[width=0.32\linewidth]{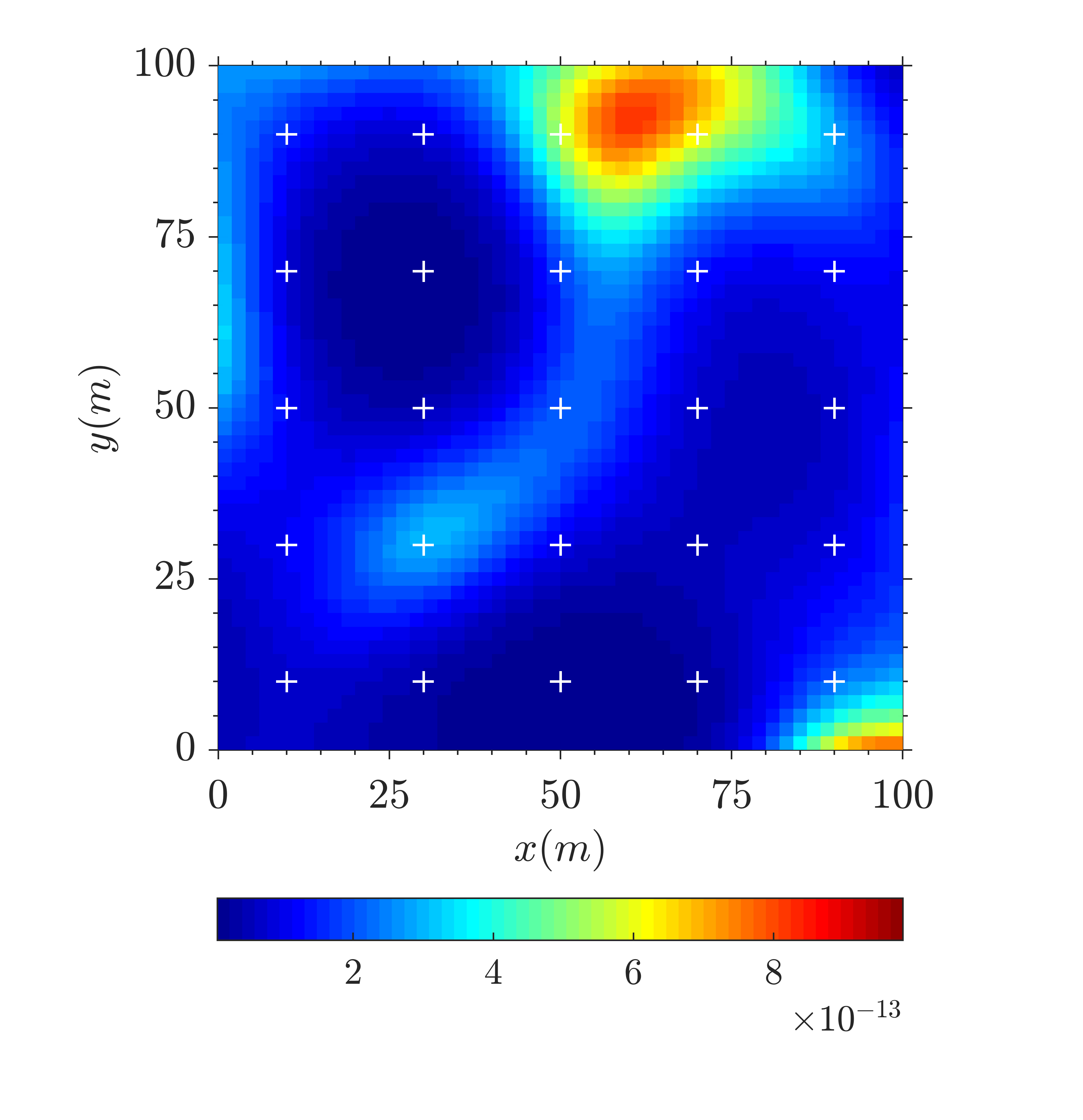}}
    \subfigure[Pressure field]{\includegraphics[width=0.32\linewidth]{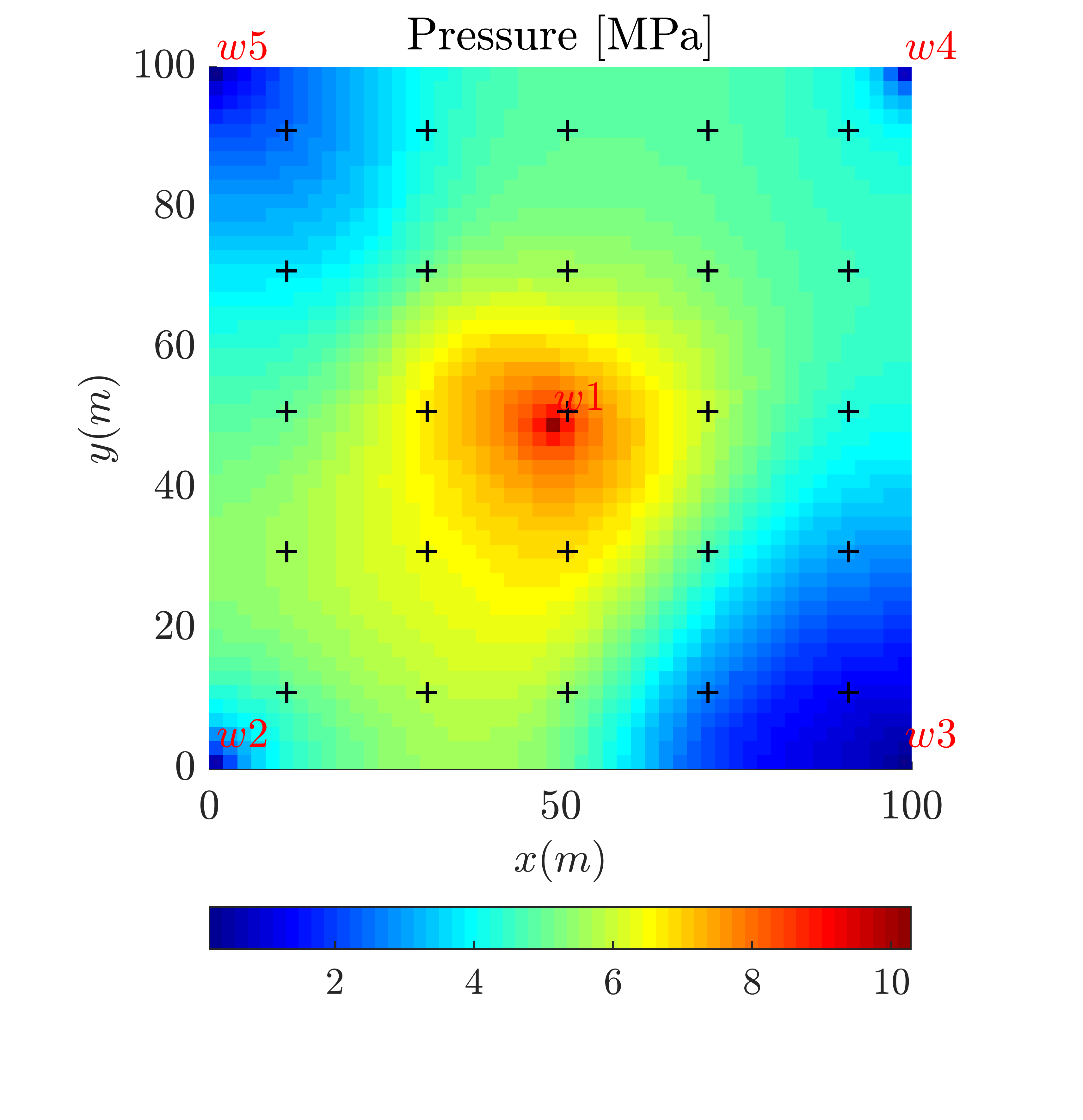}}
    \caption{Reference fields and pressure data. 
    The crosses represent the positions of the pressure sensors.}
    \label{fig:refs}
\end{figure}

\fig{fig:refs} presents the synthetic reference field $\Y$, and the corresponding permeability ($\perm$), generated using the \kl\ expansion with covariance given by \eq{eq:cov} and a correlation length $\ell=20\mathsf{m}$. Additionally, the figure includes pressure data obtained by solving \eq{eq:darcy} for the reference field.

In practical problems, due to the scarcity of information regarding the hydraulic properties of porous media, the precise description of heterogeneities is complex, and simplifying assumptions must be made. Although we are dealing with a synthetic case, we assume that we reasonably know the form of the covariance function (given here by \eq{eq:cov}) but do not know its correlation length. 

\tab{tab:exp} shows the experiments' names and the parametrization setup. In the first three experiments, we employed the \kl\ expansion with three different correlation lengths. These scenarios allow us to evaluate the impact of selecting a specific correlation length on the posterior distribution, particularly regarding convergence and acceptance rates. Here, the stochastic dimension is given by the number of modes kept in the expansion to achieve $98\%$ of the energy ($\mm$ value in \eq{eq:energy}).

In the final three experiments, we parameterized the data using the Variational Autoencoder (\VAE) method, trained on different sets of fields, as detailed in Section~\ref{sec:vaenum} (\tab{tab:exp}). In this context, we hypothesize that the Monte Carlo algorithm will identify the optimal correlation length based on the available data. In the synthetic problem addressed in this work, the correlation length used to generate the reference data is $\ell=20\mathsf{m}$. Therefore, we use different training sets that contain or do not contain fields with a correlation length equal to the reference one (see \tab{tab:data}).

\begin{table}[htbp]
\small
    \centering
    \caption{Parametrization methods used for the experiments.}
    \begin{tabular}{l|c|c}
    \hline\hline
        \textbf{Experiment} & \makecell{\textbf{Stochastic}\\[-2mm] \textbf{dimension} ($\dim$)} & \makecell[c]{\textbf{Correlation}\\[-2mm] \textbf{length} ($\ell$) $[\mathsf{m}]$}  \\ \hline
        \klen{10} & $75$ & $10$ \\
        \klen{20} & $25$ & $20$ \\
        \klen{30} & $12$ & $30$ \\
        \vaei     & $64$ & $10; 20; 30$ \\
        \vaeii    & $64$ & $15; 25; 35$ \\
        \vaeiii   & $64$ & $10; 15; 20; 25; 30; 35$ \\ \hline\hline
    \end{tabular}
    \label{tab:exp}
\end{table}

\subsubsection{Neural network training step}\label{sec:vaenum}

In this work, we used the \KL expansion to generate the sets of fields with different correlation lengths to train the neural networks. Here, we consider the complete set of eigenpairs to preserve as many details as possible.

\tab{tab:data} shows the total number of fields considered for each dataset. It is worth clarifying that \train{10-35} is not a union of \train{10-30} and \train{15-35}, i.e., the datasets are independent. Each dataset consists of $20,\!000$ fields of each correlation length. $60\%$ of the dataset is defined as the training set, $20\%$ for validation, and $20\%$ for testing, creating datasets well balanced in terms of the $\ell$ values. Furthermore, each of these three datasets will allow training a different \VAE\ neural network (\vaei, \vaeii and \vaeiii, associated with \train{10-30}, \train{15-35} and \train{10-35}, respectively).

\begin{center}
\begin{table}[!ht]
\small
\centering
\caption{Datasets information.} \label{tab:data}%
\begin{tabular}{l|c|c}
\hline\hline
\textbf{Dataset} & \makecell[c]{\textbf{Correlation}\\[-2mm] \textbf{length} ($\ell$) $[\mathsf{m}]$}  & \makecell[c]{\textbf{Total number}\\[-2mm] \textbf{of fields}}\\
\hline
\train{10-30}    & $10; 20; 30$   & $60,\!000$  \\
\train{15-35}    & $15 ; 25; 35$  & $60,\!000$  \\
\train{10-35}    & $10; 15; 20; 25; 30; 35$   & $120,\!000$ \\
\hline\hline
\end{tabular}
\end{table}
\end{center}

The same network architecture was applied to all cases. The encoder consists of three convolutional layers followed by a dense layer. Each convolutional layer utilizes four filters with a kernel size of $5$, incorporates batch normalization, and employs the ReLU activation function. The second convolutional layer has a stride of $2$. The dense layer contains $1,\!024$ neurons and uses the ReLU activation function. Additionally, we use a latent dimension of $64$. The network architecture and other parameters were defined in preliminary studies, seeking the smallest possible latent dimension.

\begin{figure}[htbp]
\centering
\subfigure[Reconstruction loss]{\includegraphics[width=0.32\textwidth]{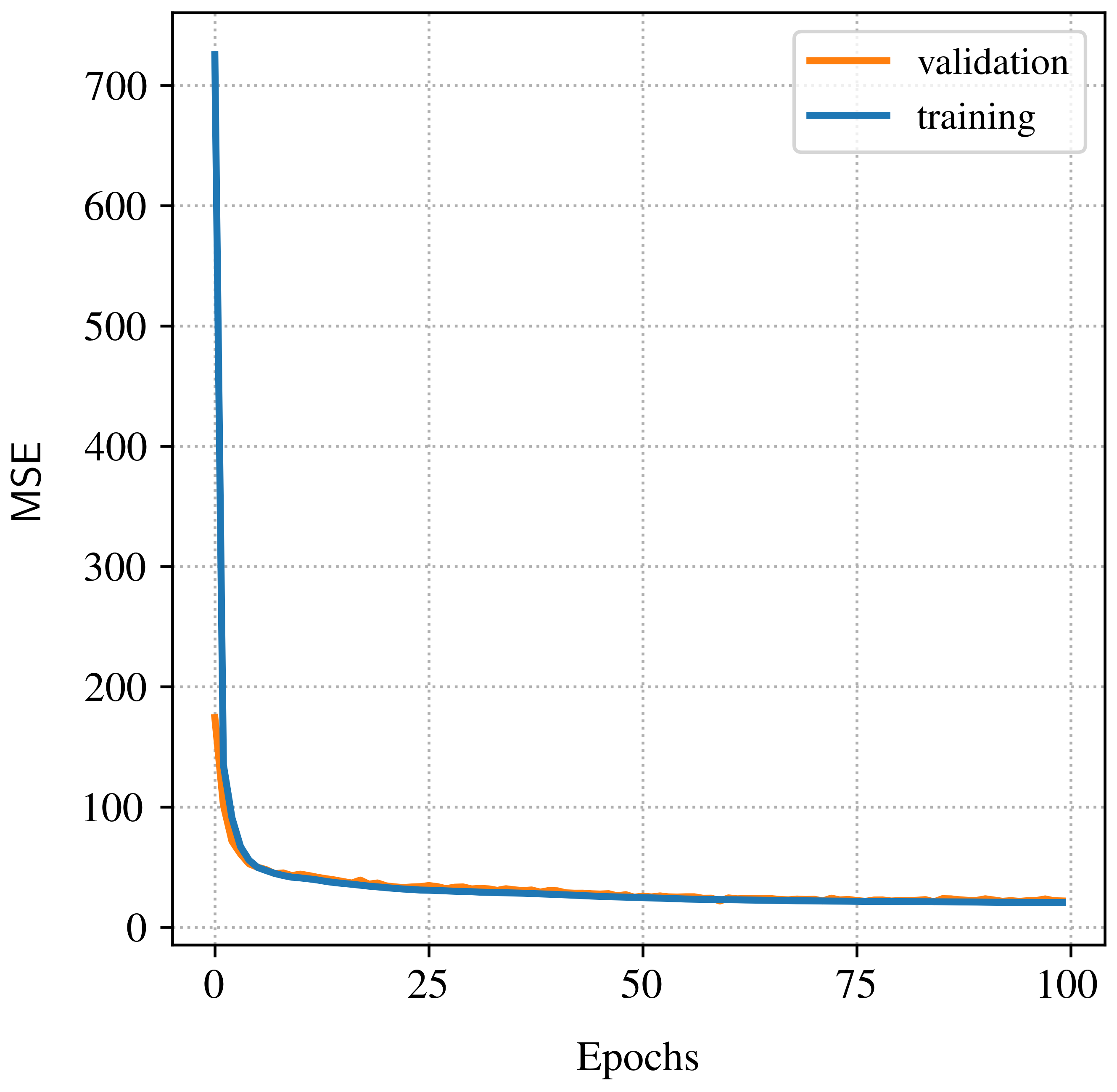}} 
\subfigure[\kldiv divergence]{\includegraphics[width=0.32\textwidth]{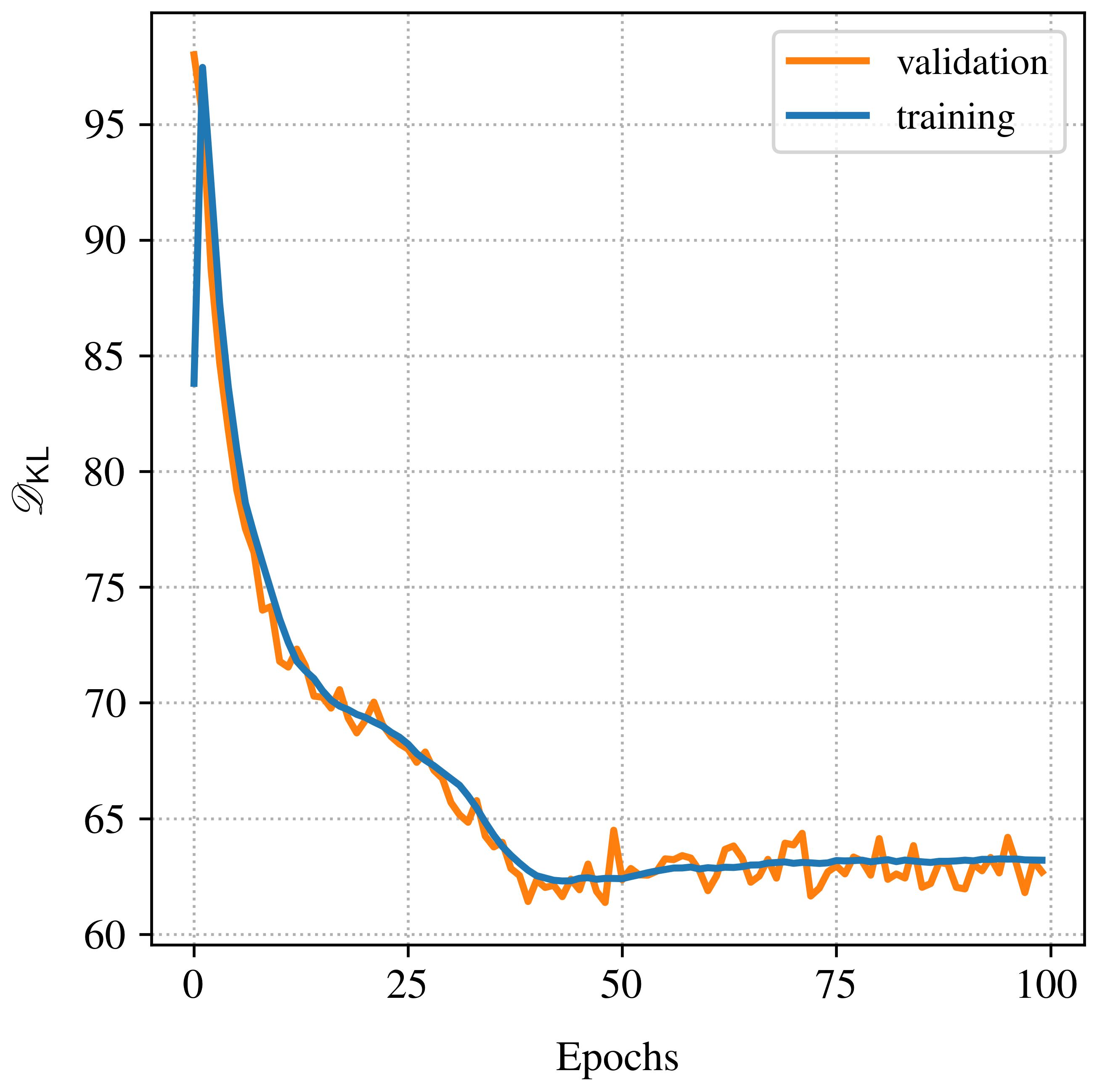}} 
\subfigure[Total loss]{\includegraphics[width=0.32\textwidth]{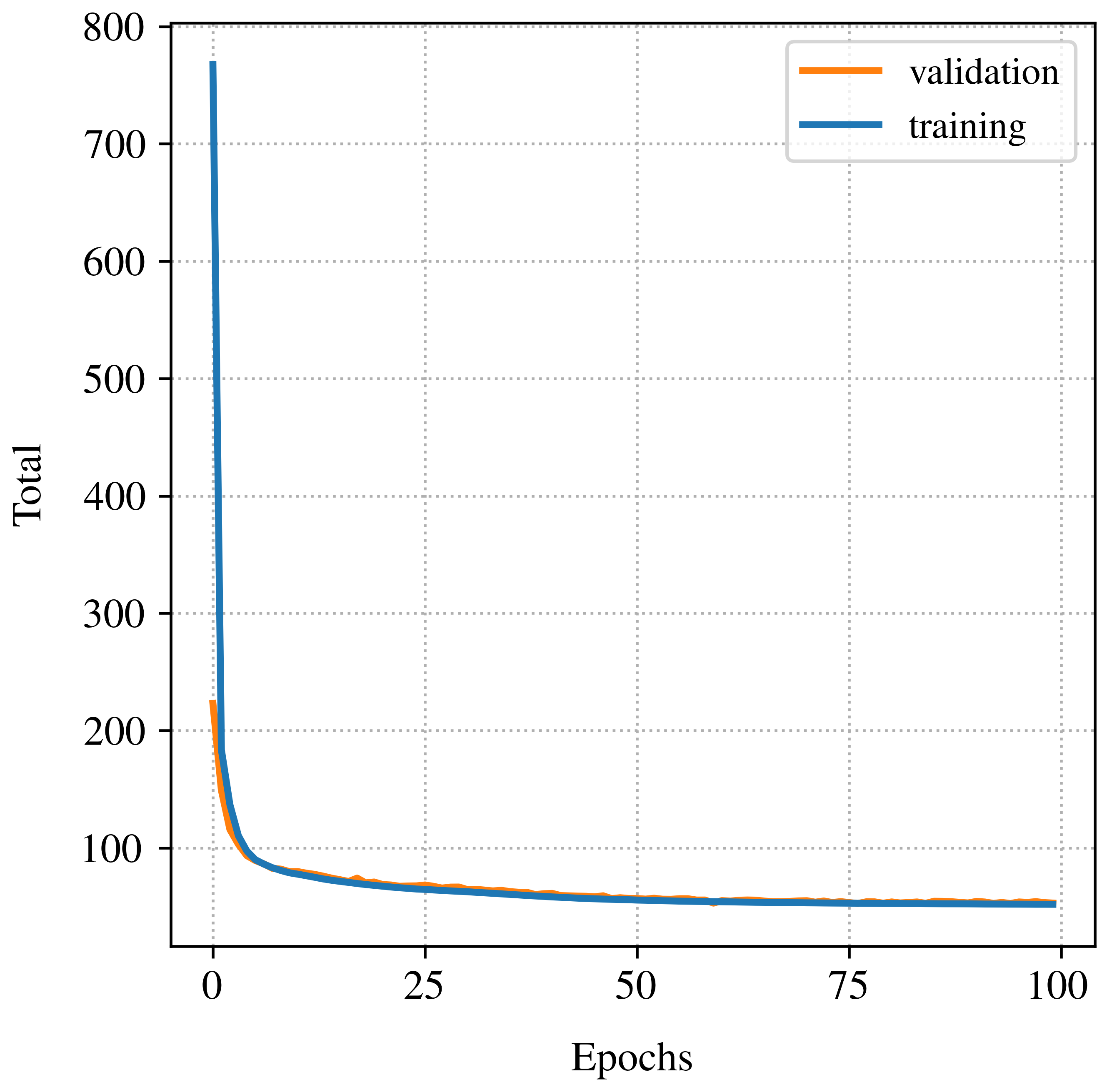}} \\
\caption{Convergence metrics for the \vaei training.}
\label{fig:vae1}
\end{figure}

\tab{tab:enc_arch} summarizes the encoder architecture. The decoder mirrors the encoders. We used a batch size of $25$ and trained for $100$ epochs. The optimization process utilized the Adam optimizer with a $10^{-4}$ learning rate. Lastly, we set $\beta=0.5$ in the loss function to improve the reconstruction quality. The reconstruction loss ($\lmse$), $\mathcal{KL}$-divergence ($\dkl$), and total losses are illustrated in \figsto{fig:vae1}{fig:vae3}. The training converged successfully for all three networks. The validation sets closely followed the training sets due to the high representational power of the training data. 

As previously mentioned, the acceptance rate and convergence of \mcmc\ algorithms are highly sensitive to the stochastic dimension of the problem, and therefore, this is a critical point in this work. The latent dimension of $64$ is within the range of stochastic dimensions $[12, 75]$ for the cases in which the \kle\ method was used (\tab{tab:exp}). Thus, we consider that the dimension reduction was significant.


\begin{table}[!htbp]
\tiny
\centering
\caption{Model summary of the VAEs encoder.} \label{tab:enc_arch}%
\begin{tabular}{c|c|c|c}
\hline \hline
\textbf{Layer (type)} & \textbf{Output Shape} & \textbf{Param \#} & \textbf{Connected to} \\ 
\hline
input (InputLayer) & (None, 50, 50, 1) & 0 & - \\ 
conv2d (Conv2D) & (None, 50, 50, 4) & 100 & input \\ 
batch\_norm (BatchNormalization) & (None, 50, 50, 4) & 16 & conv2d \\ 
activation (Activation) & (None, 50, 50, 4) & 0 & batch\_norm \\ 
conv2d\_1 (Conv2D) & (None, 25, 25, 4) & 400 & activation \\ 
batch\_norm\_1 (BatchNormalization) & (None, 50, 50, 4) & 16 & conv2d \\ 
activation\_1 (Activation) & (None, 25, 25, 4) & 0 & batch\_norm\_1 \\ 
conv2d\_2 (Conv2D) & (None, 25, 25, 4) & 400 & activation\_1 \\ 
batch\_norm\_2 (BatchNormalization) & (None, 50, 50, 4) & 16 & conv2d \\ 
activation\_2 (Activation) & (None, 25, 25, 4) & 0 & batch\_norm\_2 \\ 
flatten (Flatten) & (None, 2500) & 0 & activation\_2 \\ 
dense (Dense) & (None, 1024) & 2,560,000 & flatten \\ 
activation\_3 (Activation) & (None, 1024) & 0 & dense \\ 
z\_mean (Dense) & (None, 64) & 65,600 & activation\_3 \\ 
z\_log\_var (Dense) & (None, 64) & 65,600 & activation\_3 \\ 
z (Sampling) & (None, 64) & 0 & z\_mean, z\_log\_var \\ 
\hline \hline
\end{tabular}
\end{table}

\begin{figure}[htbp]
\centering
\subfigure[\vaei]{\includegraphics[width=0.32\textwidth]{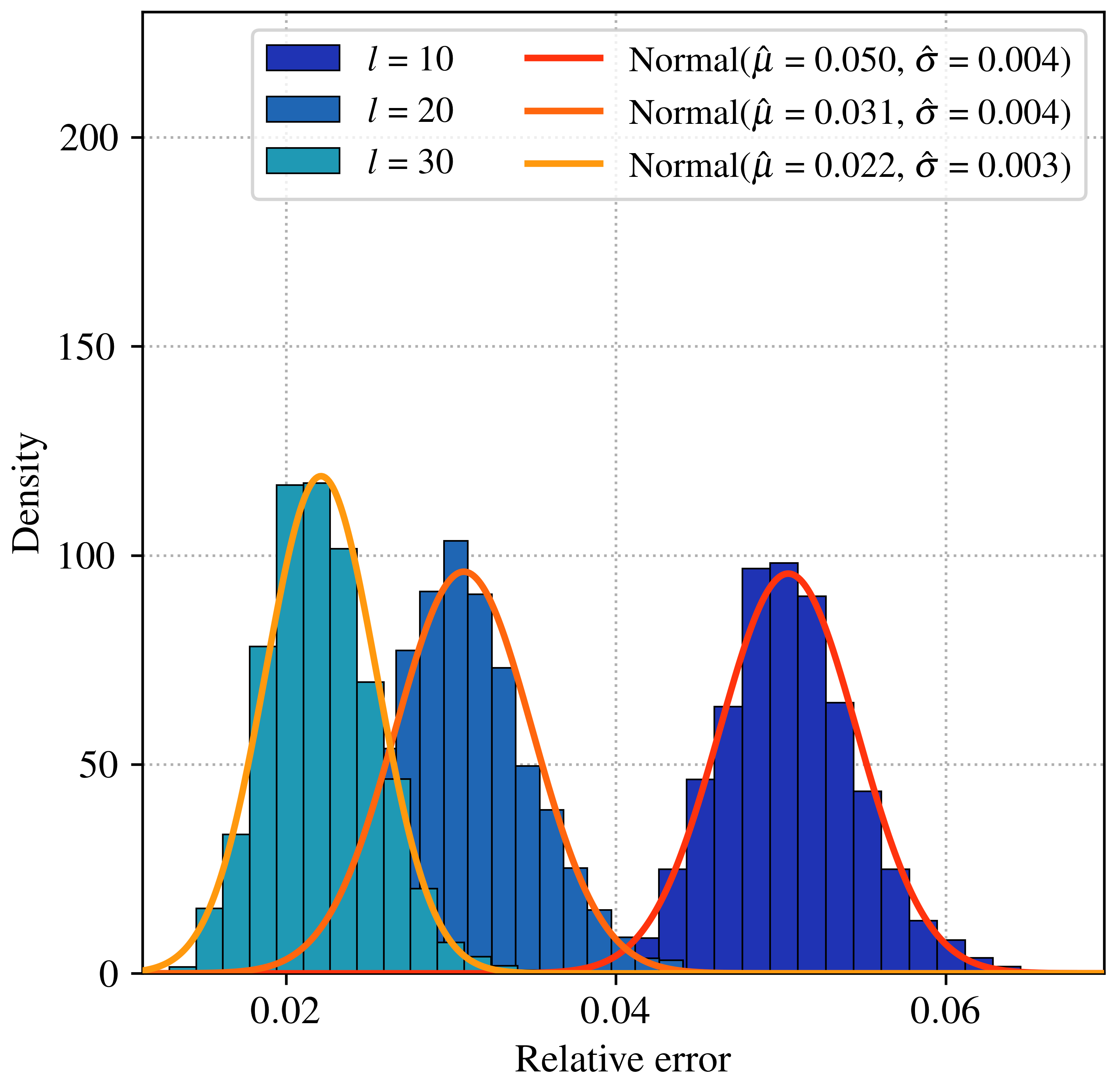}}  
\subfigure[\vaeii]{\includegraphics[width=0.32\textwidth]{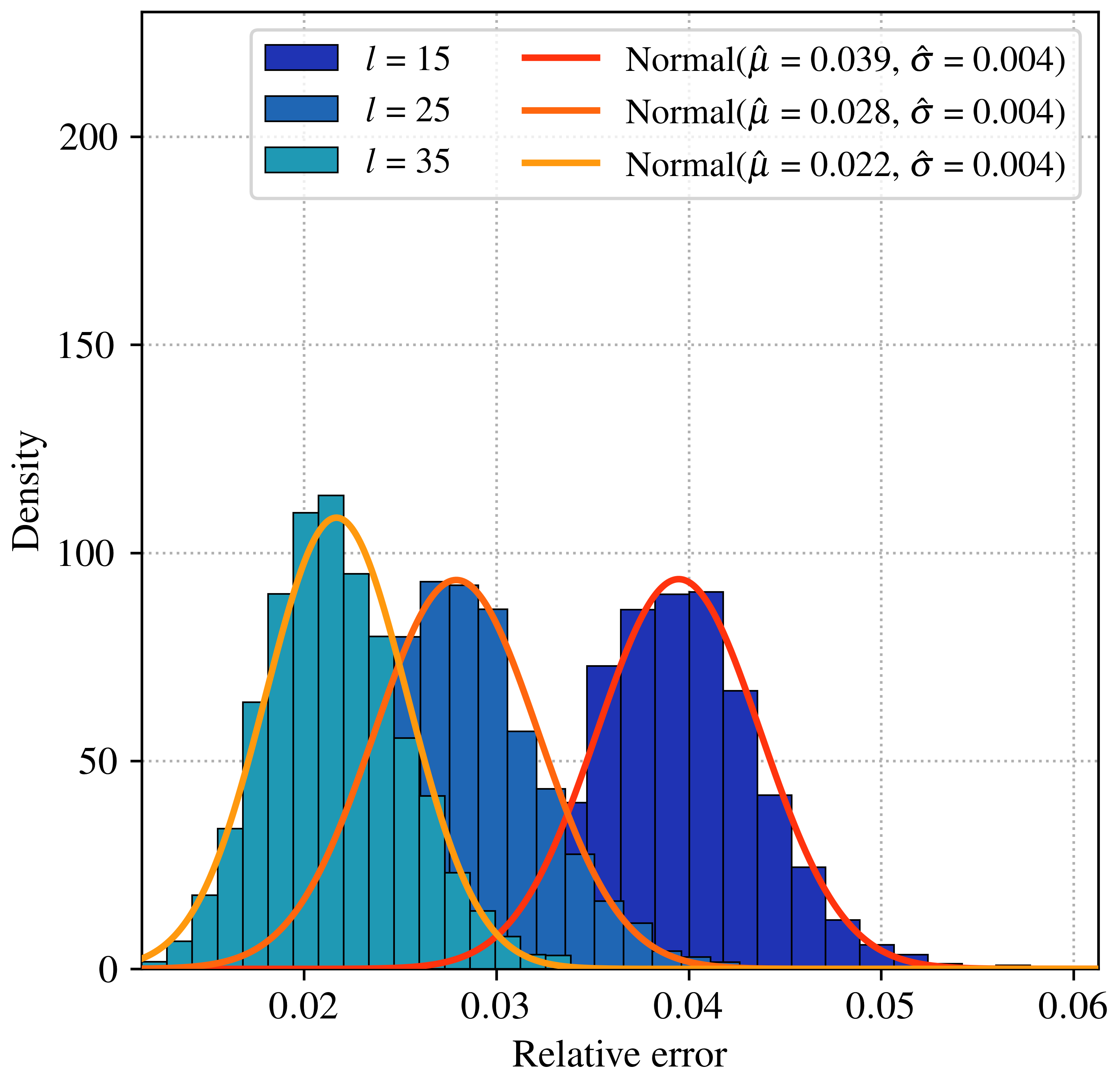}}  
\subfigure[\vaeiii]{\includegraphics[width=0.32\textwidth]{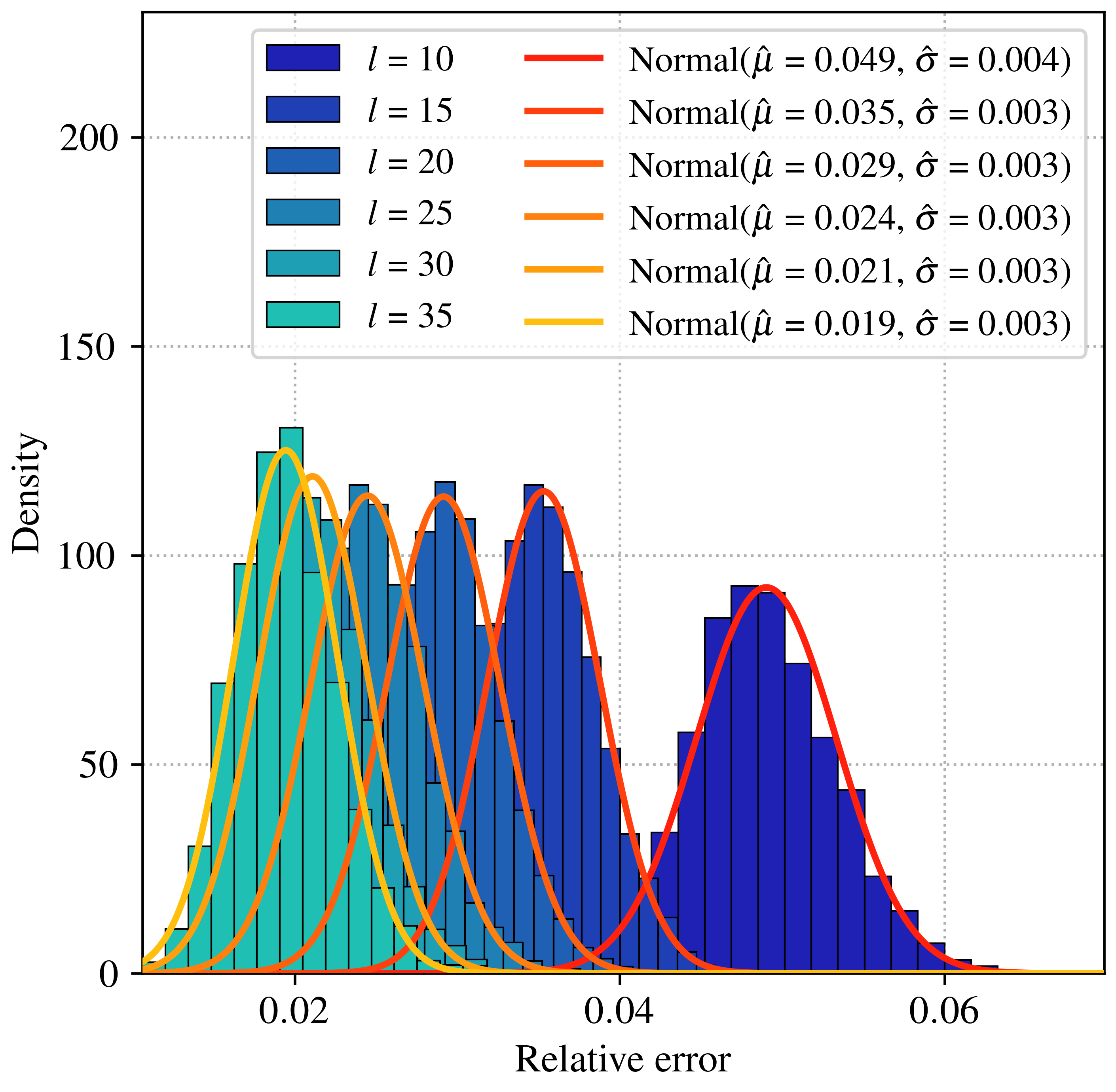}} \\
\caption{Relative error histograms for each testing set, organized by $\ell$ value.}
\label{fig:VAEhisterrors}
\end{figure}

For a general overview of the testing results, \fig{fig:VAEhisterrors} presents the relative error histograms for each network, using all $4,\!000$ images corresponding to each value of $\ell$ in its respective testing dataset. Although the quadratic error was defined for measuring reconstruction loss, the relative error is more straightforward for comparative analysis. Larger errors are associated with smaller correlation lengths, as the fields with smaller $\ell$ values are less smooth. Consequently, these fields require a larger latent dimension, similar to the behavior observed with the \kle\ strategy. Overall, the results align with our expectations, and the error distributions exhibit a Gaussian distribution pattern.

\begin{figure}[htbp]
\centering
\subfigure[Reconstruction loss]{\includegraphics[width=0.32\textwidth]{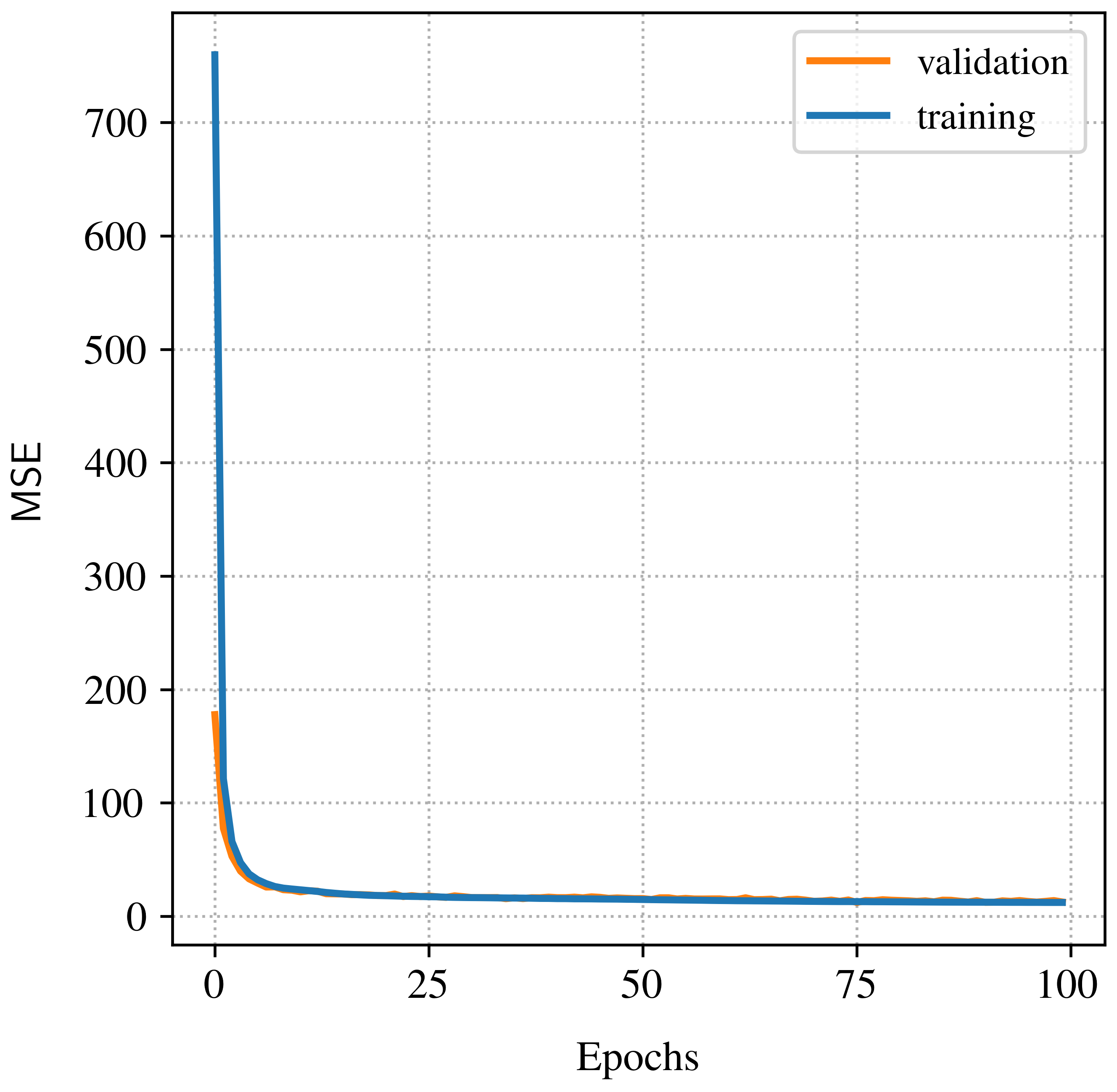}} 
\subfigure[\kldiv divergence]{\includegraphics[width=0.32\textwidth]{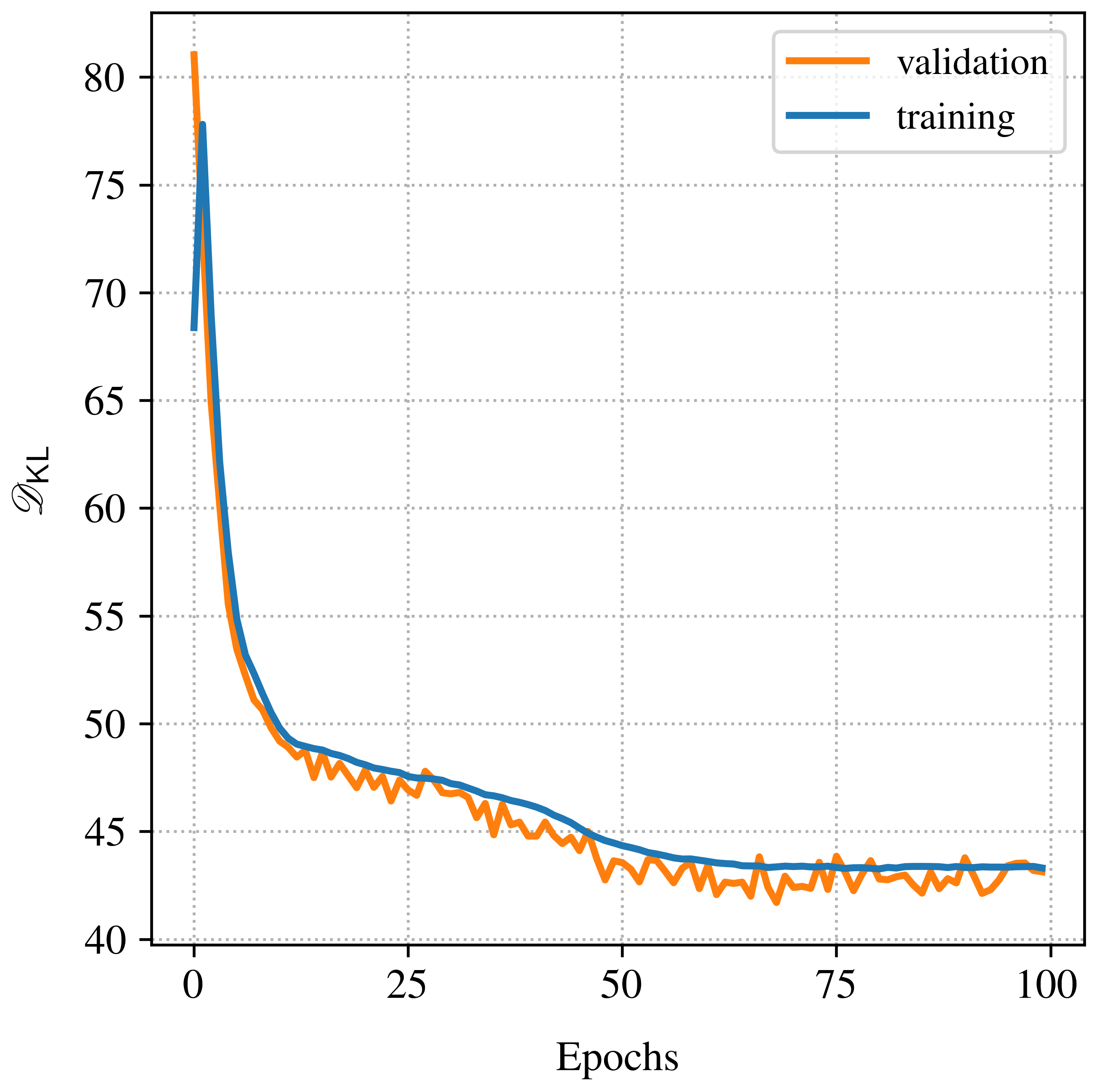}} 
\subfigure[Total loss]{\includegraphics[width=0.32\textwidth]{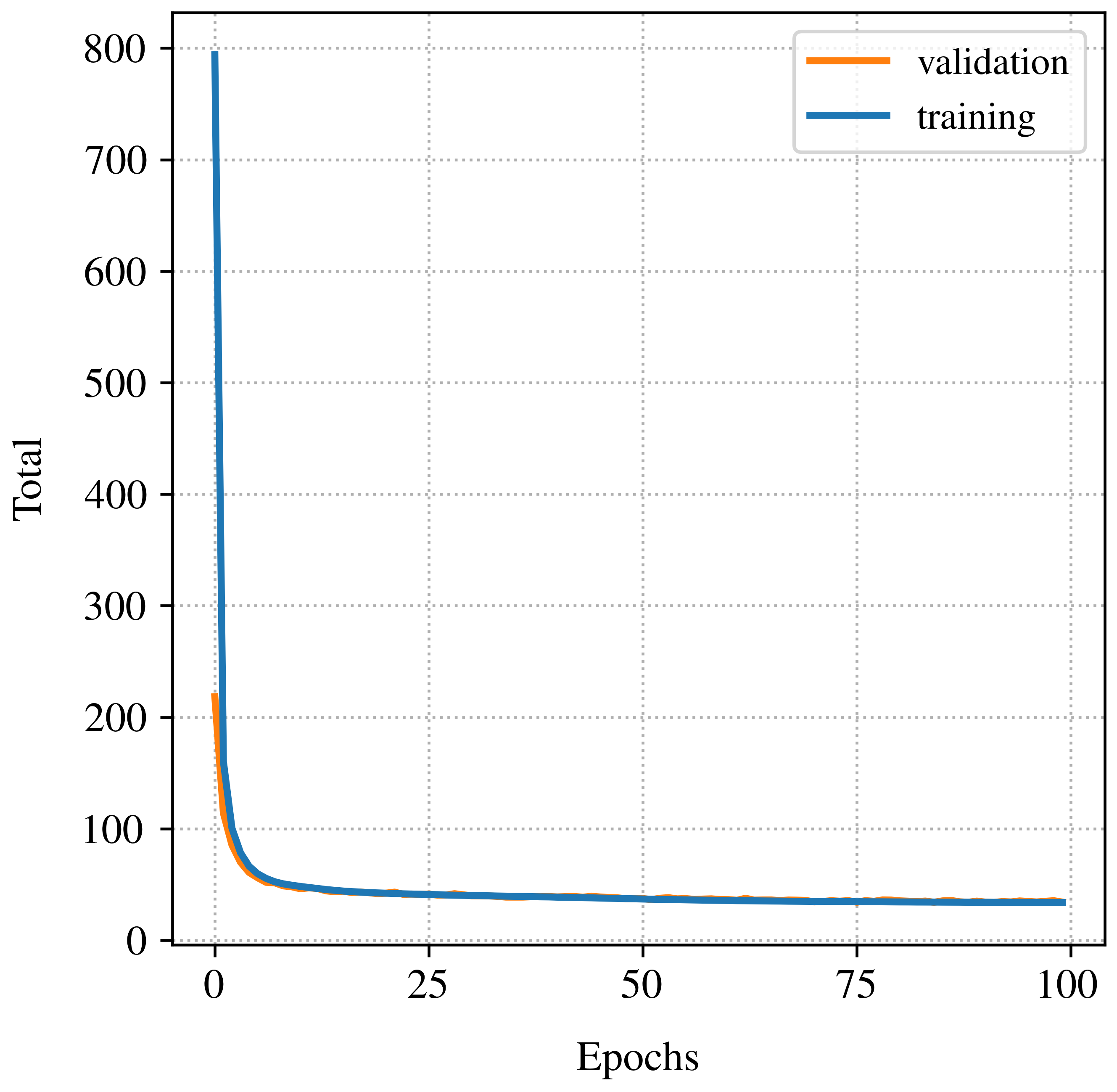}} \\
\caption{Convergence metrics for the \vaeii training.}
\label{fig:vae2}
\end{figure}

\begin{figure}[htbp]
\centering
\subfigure[Reconstruction loss]{\includegraphics[width=0.32\textwidth]{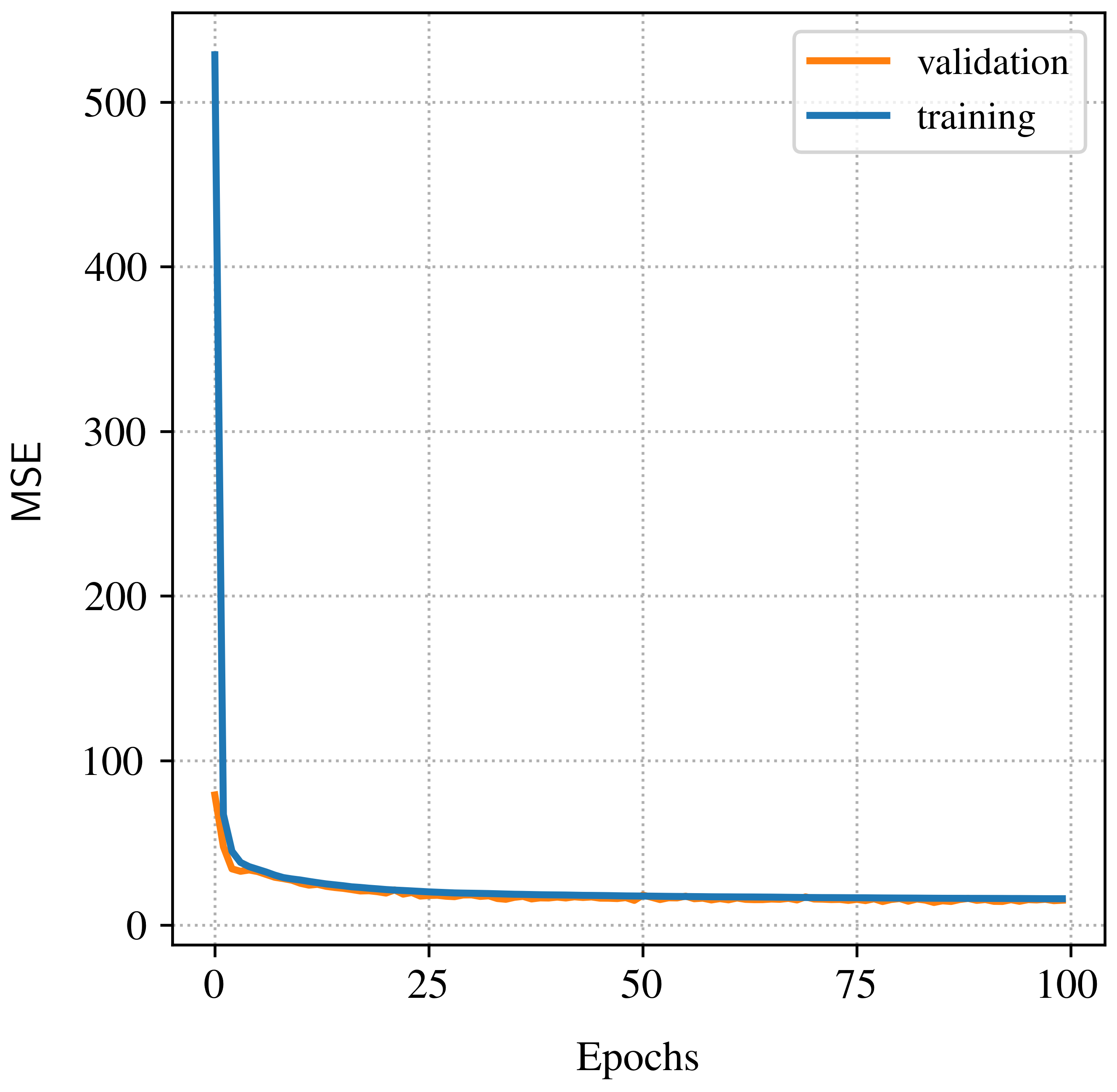}} 
\subfigure[\kldiv divergence]{\includegraphics[width=0.32\textwidth]{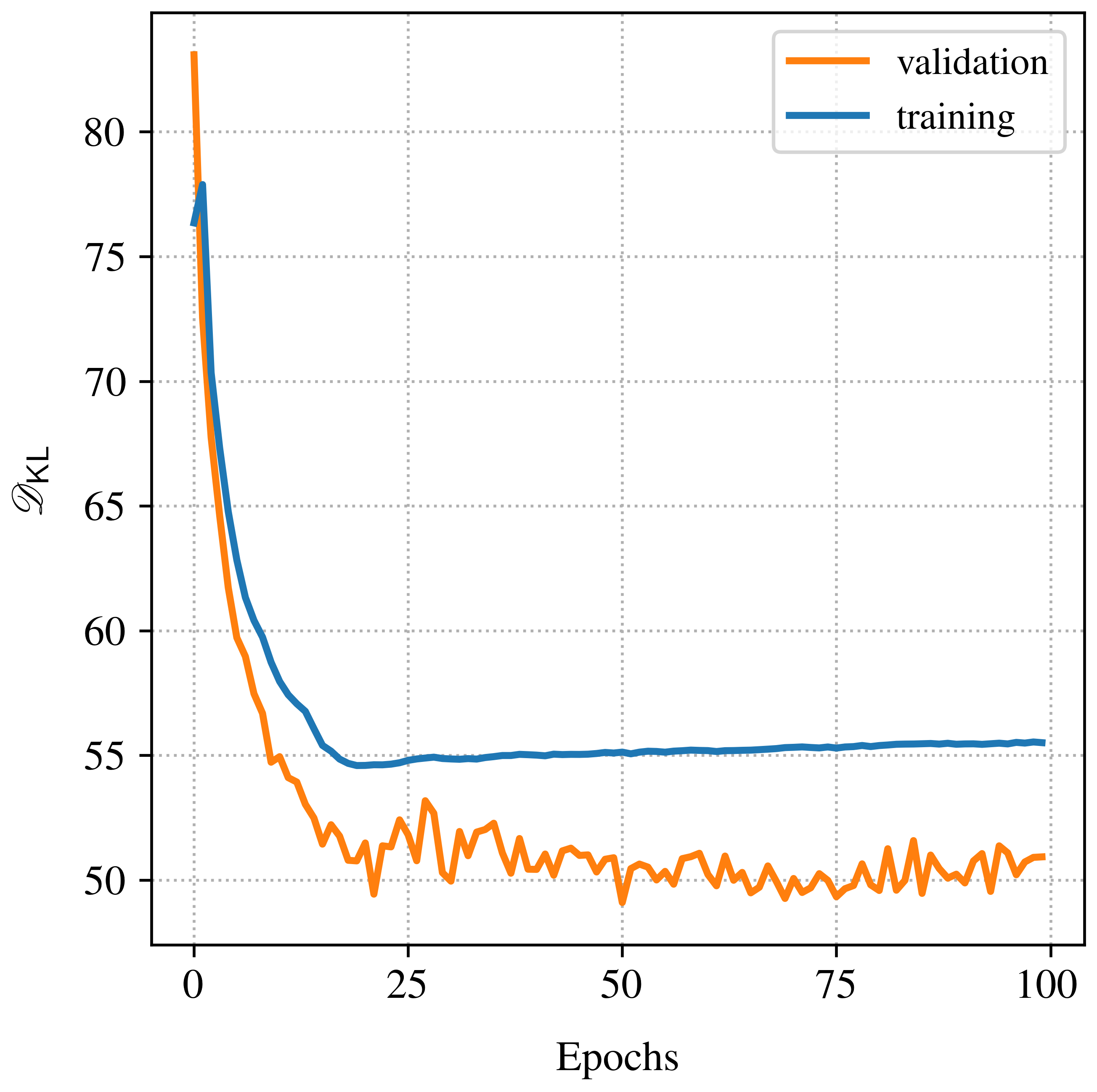}} 
\subfigure[Total loss]{\includegraphics[width=0.32\textwidth]{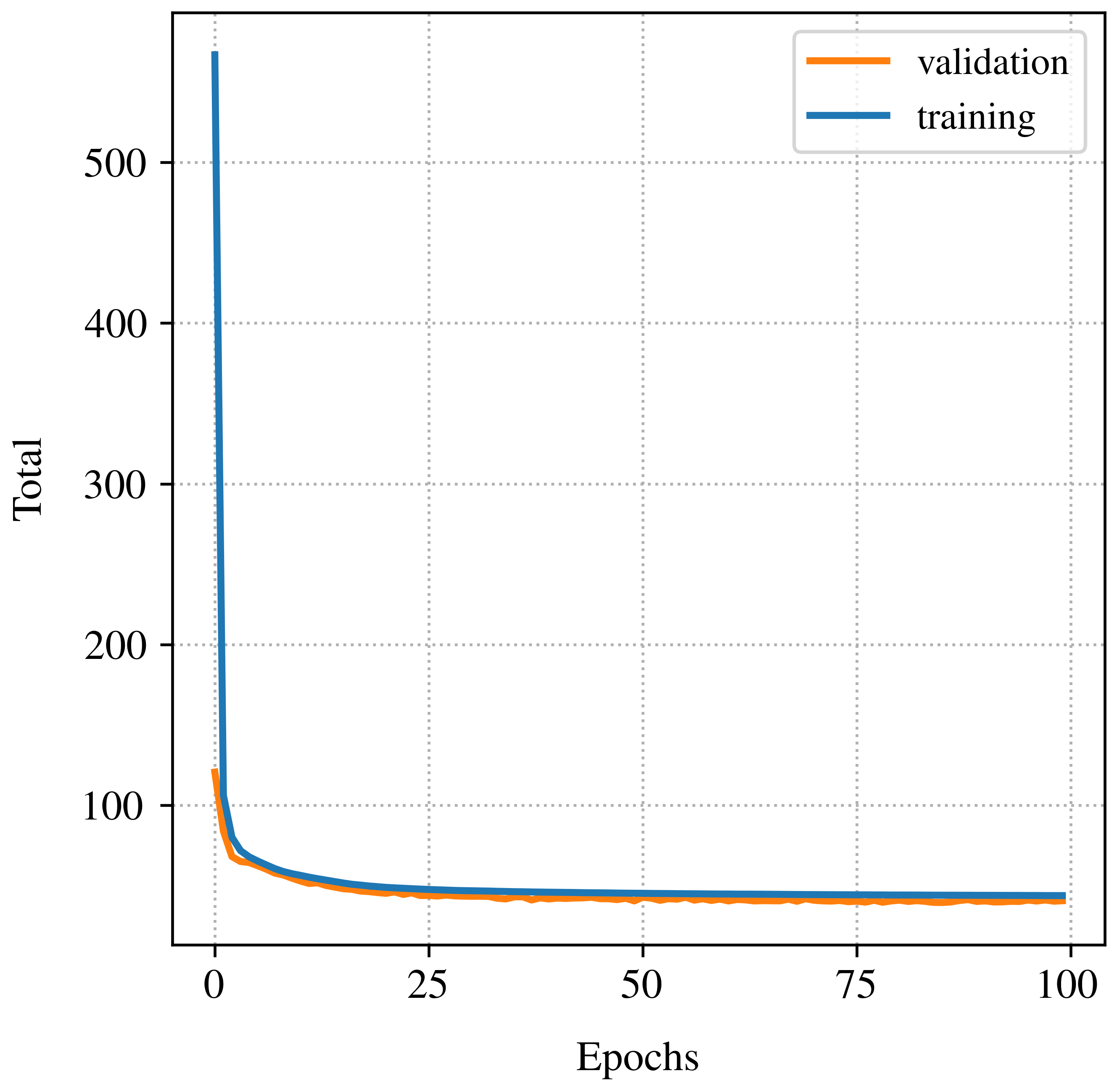}} \\
\caption{Convergence metrics for the \vaeiii training.}
\label{fig:vae3}
\end{figure}

\clearpage
\subsection{\mcmc\ simulations}

Each experiment involved $16$ independent parallel chains, with $300,\!000$ iterations performed for each chain. The initial $10,\!000$ states from each chain were removed (burn-in). 

\begin{figure}[htbp]
    \centering
    \subfigure[\klen{10}]{\includegraphics[width=0.32\linewidth]{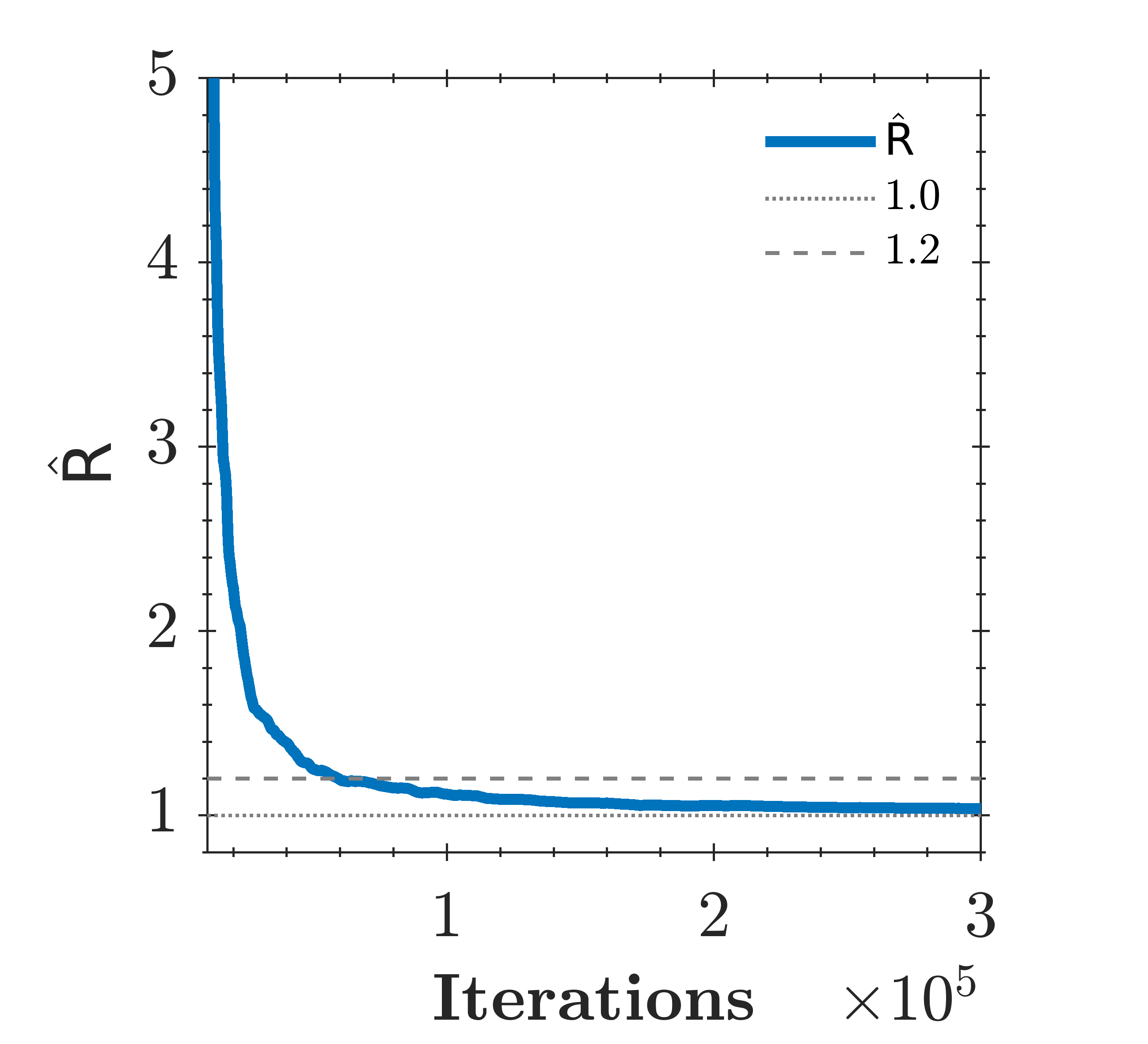}}
    \subfigure[\klen{20}]{\includegraphics[width=0.32\linewidth]{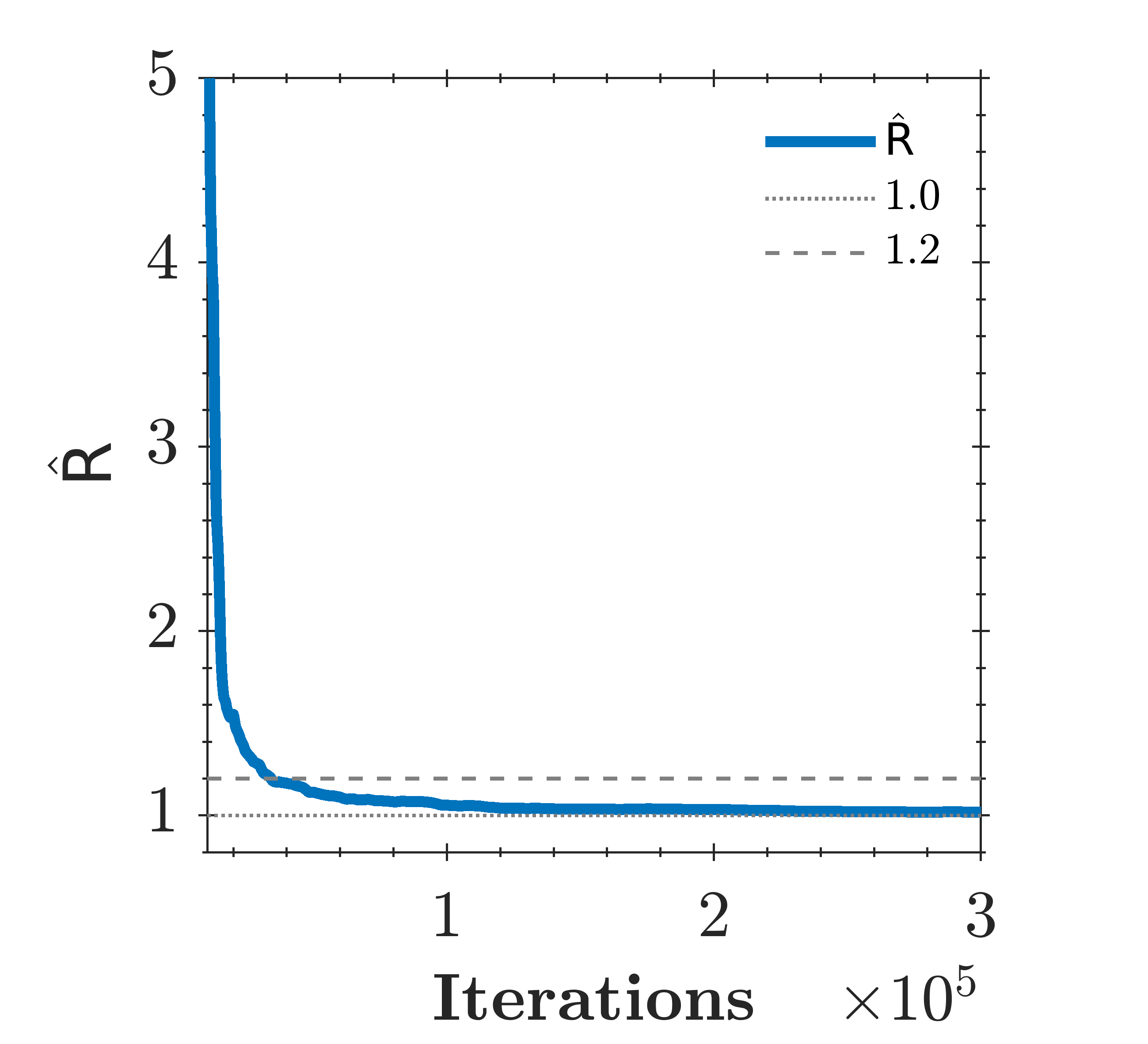}}
    \subfigure[\klen{30}]{\includegraphics[width=0.32\linewidth]{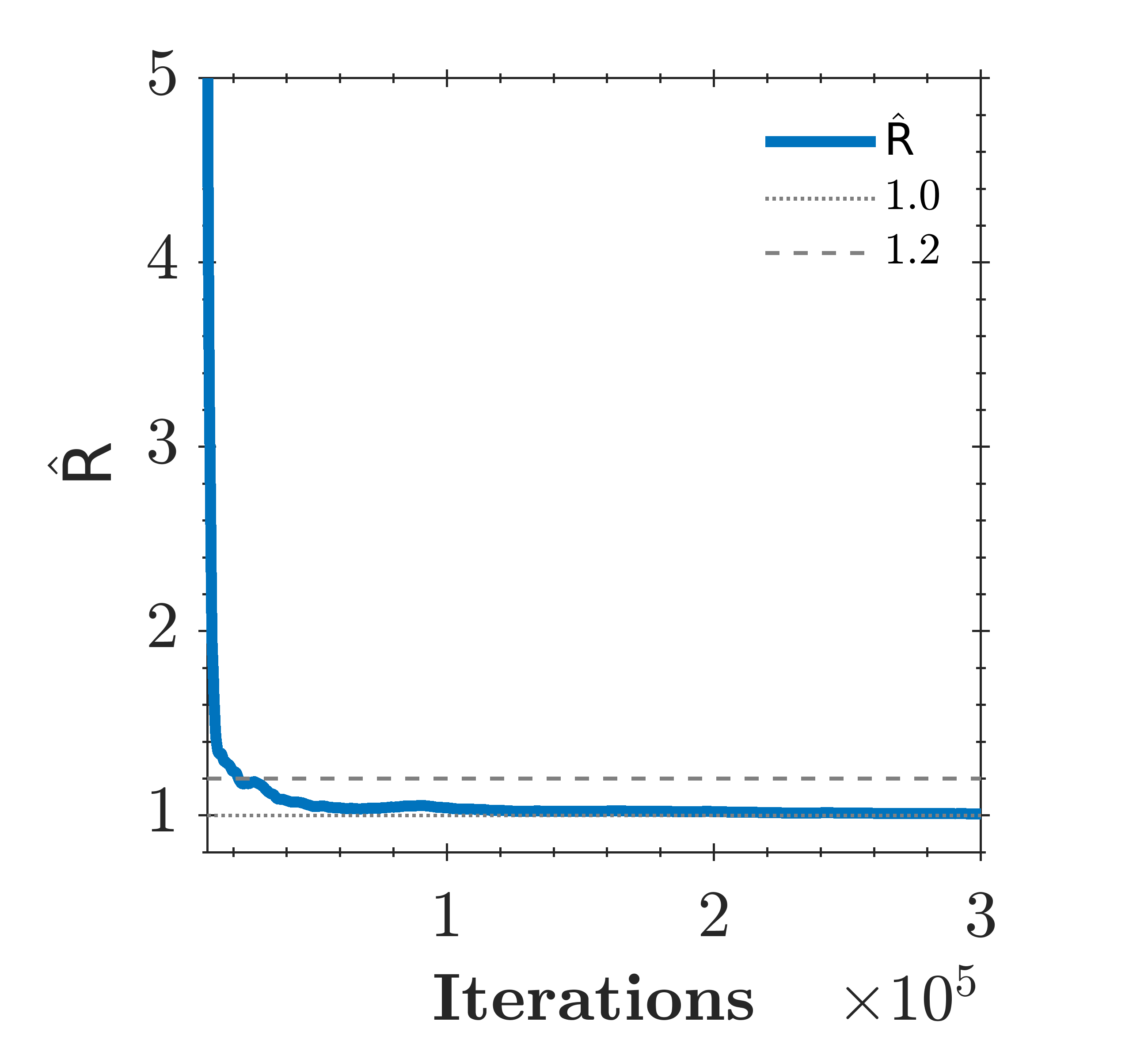}} \\
    \subfigure[\vaei]{\includegraphics[width=0.32\linewidth]{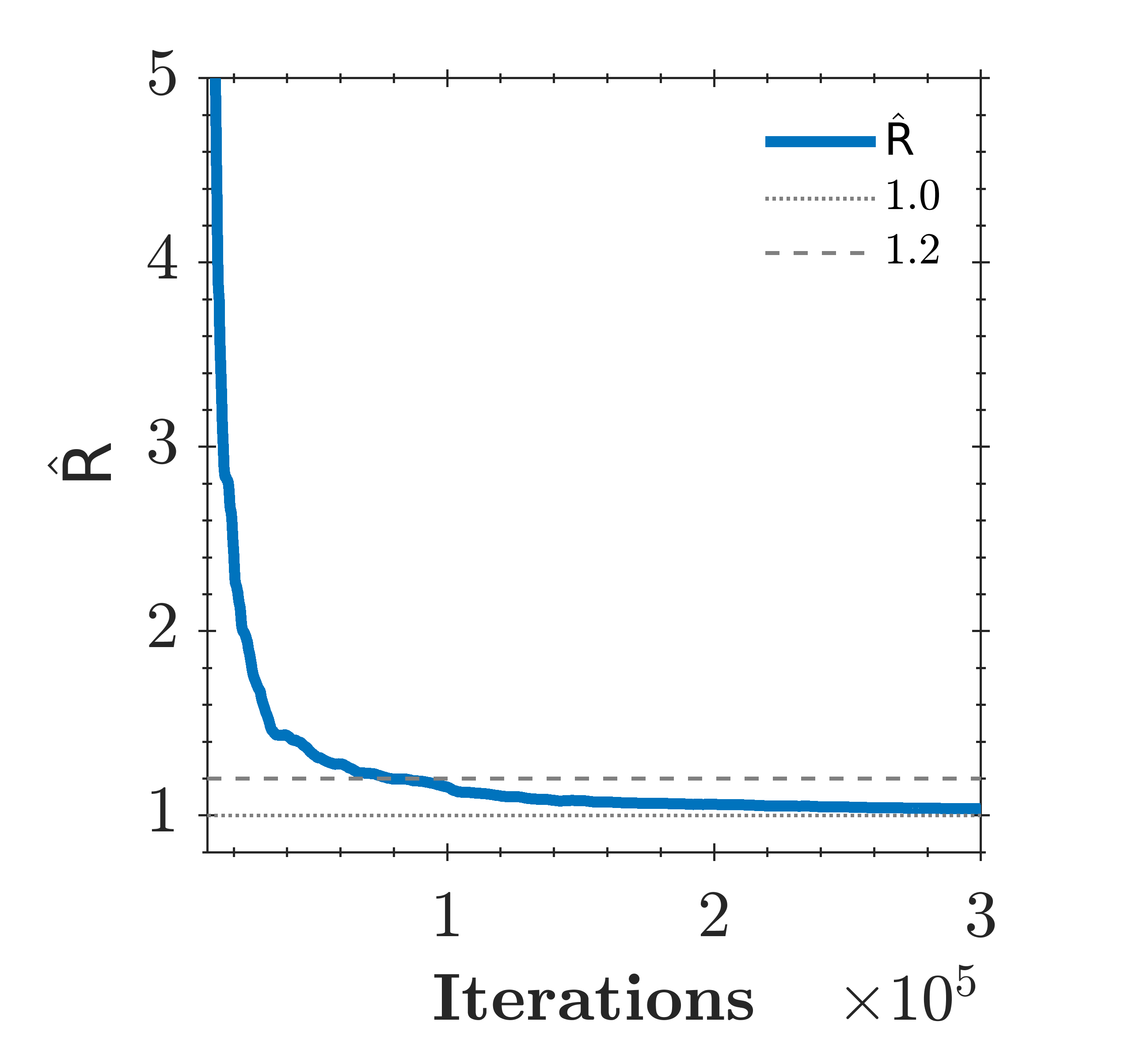}}
    \subfigure[\vaeii]{\includegraphics[width=0.32\linewidth]{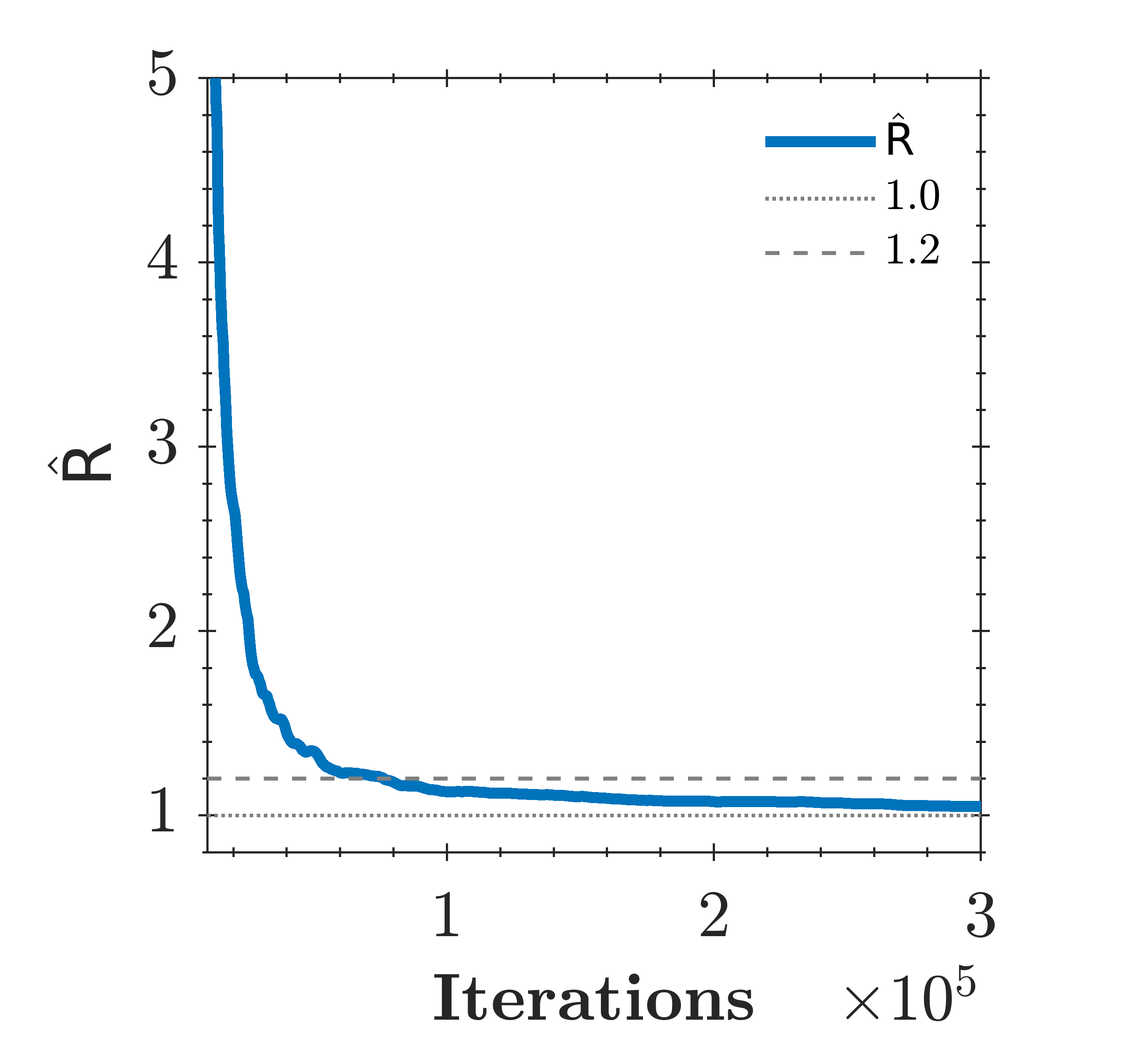}}
    \subfigure[\vaeiii]{\includegraphics[width=0.32\linewidth]{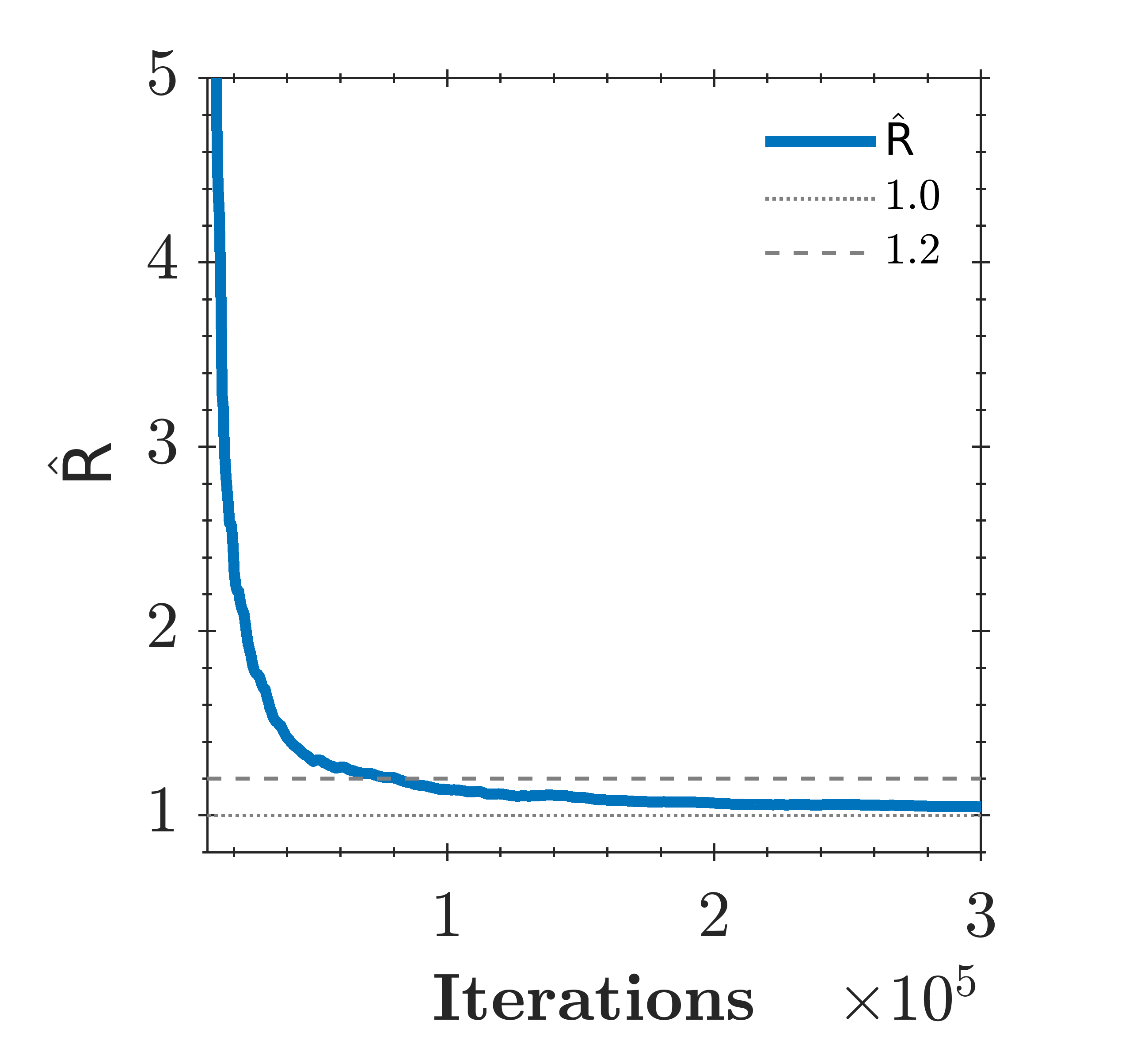}} \\
    \caption{\MPSRF\ results for each experiment.}
    \label{fig:R}
\end{figure}

When using \mcmc\ methods, we must first check the convergence of the multiple chains. In this study, we assess the convergence using the multivariate potential scale reduction factor (\MPSRF\ or $\psrf$) proposed by \citet{GelmanRubin1992}. As the authors recommended, we establish that convergence was achieved when $\psrf < 1.2$. \fig{fig:R} plots the estimated \MPSRF\ for each experiment. The \tab{tab:AR} displays the number of iterations until $\psrf$ falls below $1.2$ for our study. As expected, the higher the stochastic dimension (\tab{tab:data}), the higher the number of iterations required until convergence. Note that all cases reached convergence before $81,\!000$ iterations. Thus, to study the posterior distribution, we use the last $50,\!000$ states of each chain, totaling $800,\!000$ states per experiment, from which we randomly select $\Np = 10,\!000$ samples for the remaining studies. The \klen{20} experiment will be used as the gold standard solution for comparison purposes since it used the correct correlation length in your prior distribution.

After the convergence, the acceptance rate ($\AR$) can be used to evaluate the efficiency of the algorithm \citep{roberts2001}. It is defined as the probability (in stationarity) that the proposed move of a chain is accepted. In this work, we compute the acceptance rate as follows:
\begin{equation}
    \AR = \dfrac{\na}{\nruns} \times 100,
    \label{eq:AR}
\end{equation}
\noindent where $\na$ is the number of accepted moves in $\nruns$ movements for each chain after eliminating the burn-in period. Furthermore, we define the mean acceptance rate ($\eAR$) for each experiment as the average of the $\AR$ of the multiple chains that compose the experiment. 

\tab{tab:AR} shows the estimated mean acceptance rate for each experiment. None of the cases stood out in this aspect. As we will see later, although \klen{10} obtained the highest acceptance rate, this did not impact the quality of the posterior distribution of the fields obtained compared to the \klen{20} case.

\begin{figure}[htbp]
    \centering
    \subfigure[\klen{10}]{\includegraphics[width=0.32\linewidth]{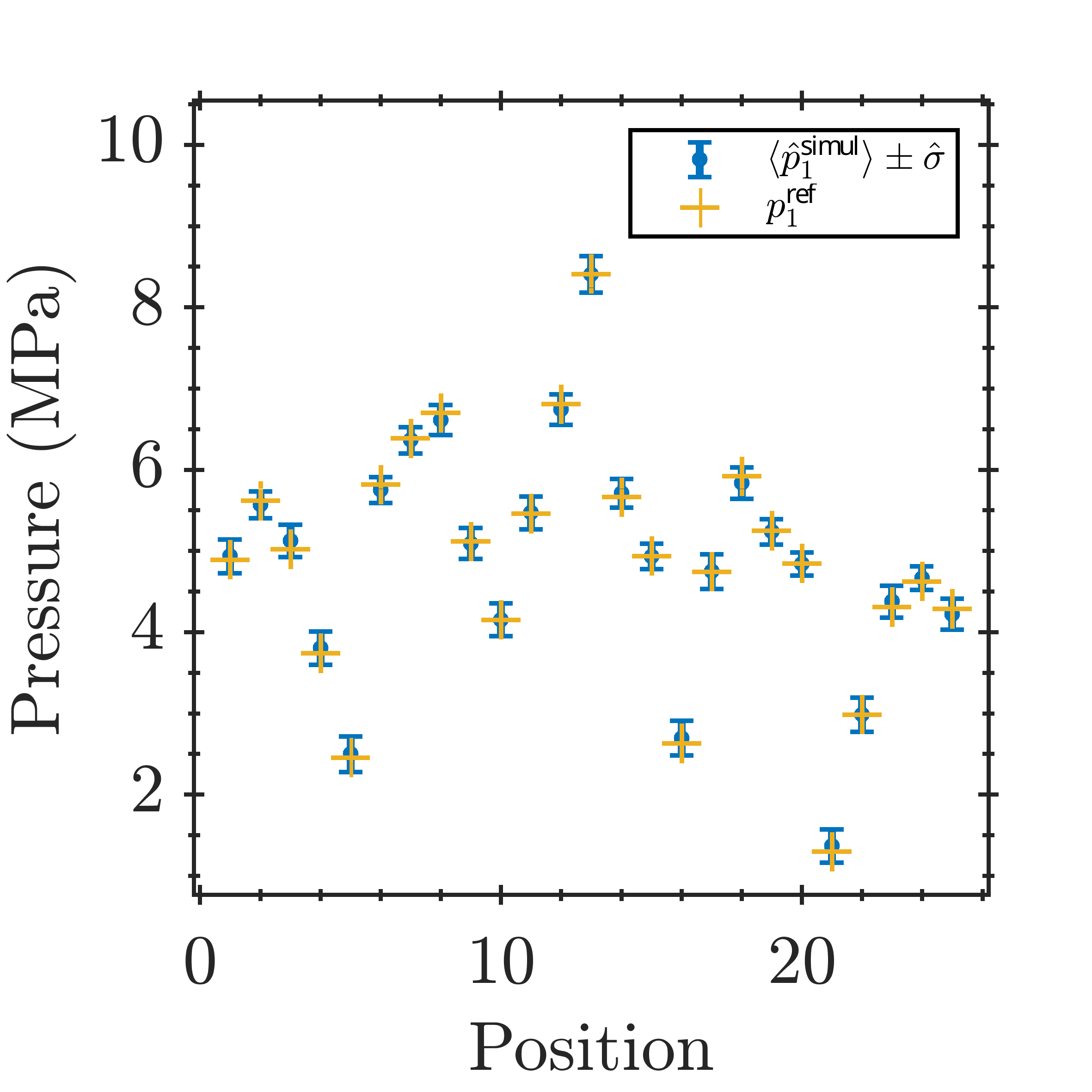}}
    \subfigure[\klen{20}]{\includegraphics[width=0.32\linewidth]{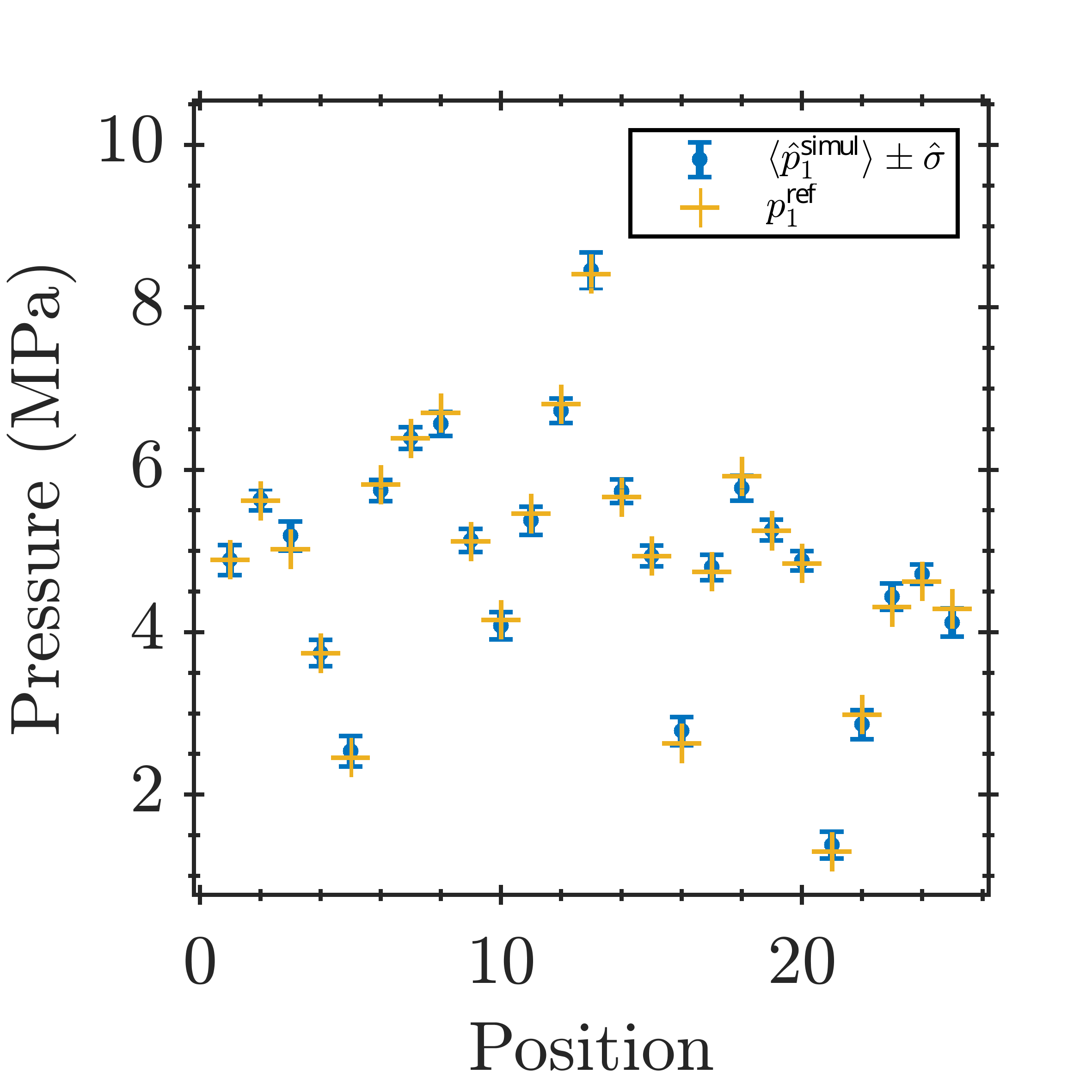}}
    \subfigure[\klen{30}]{\includegraphics[width=0.32\linewidth]{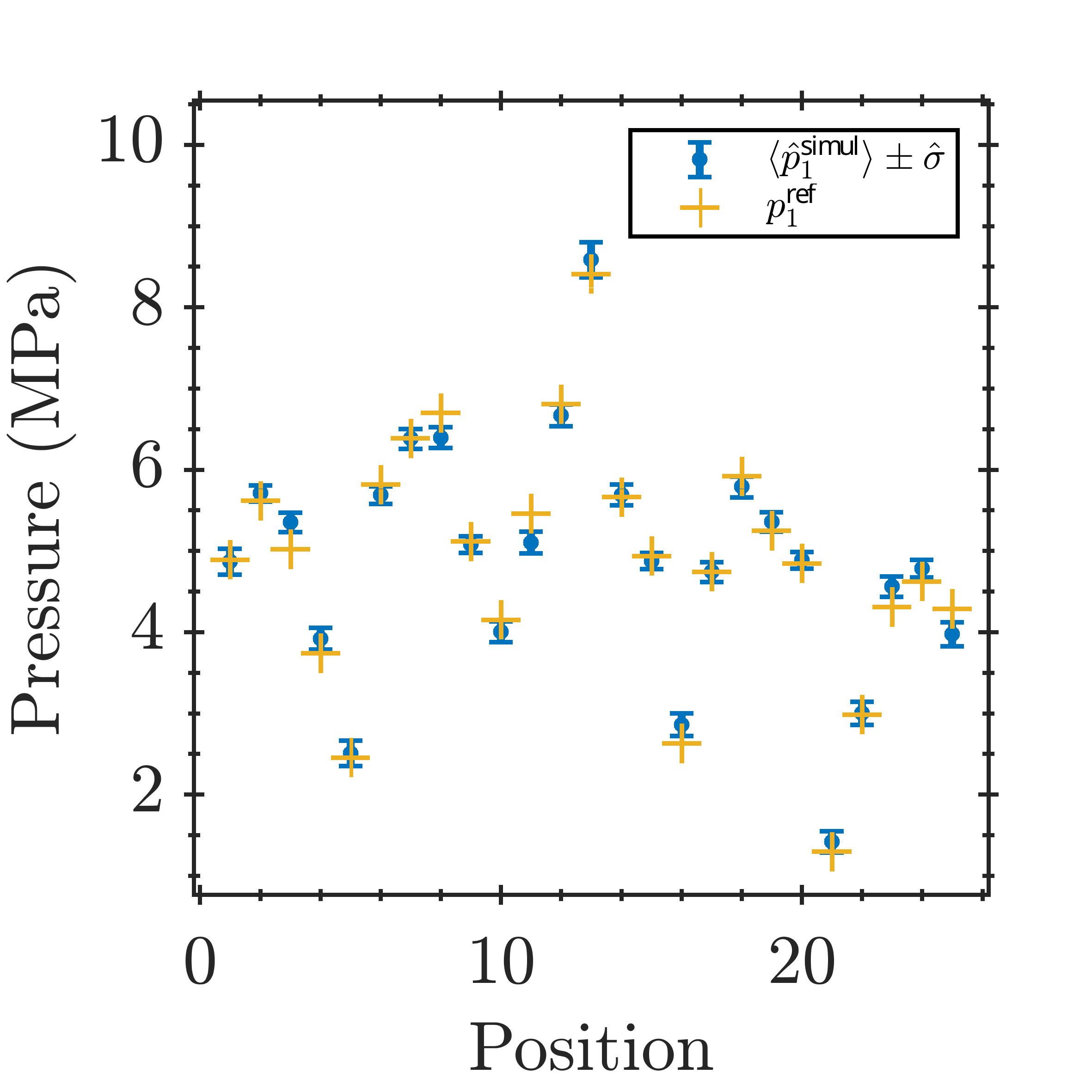}} \\
    \subfigure[\vaei]{\includegraphics[width=0.32\linewidth]{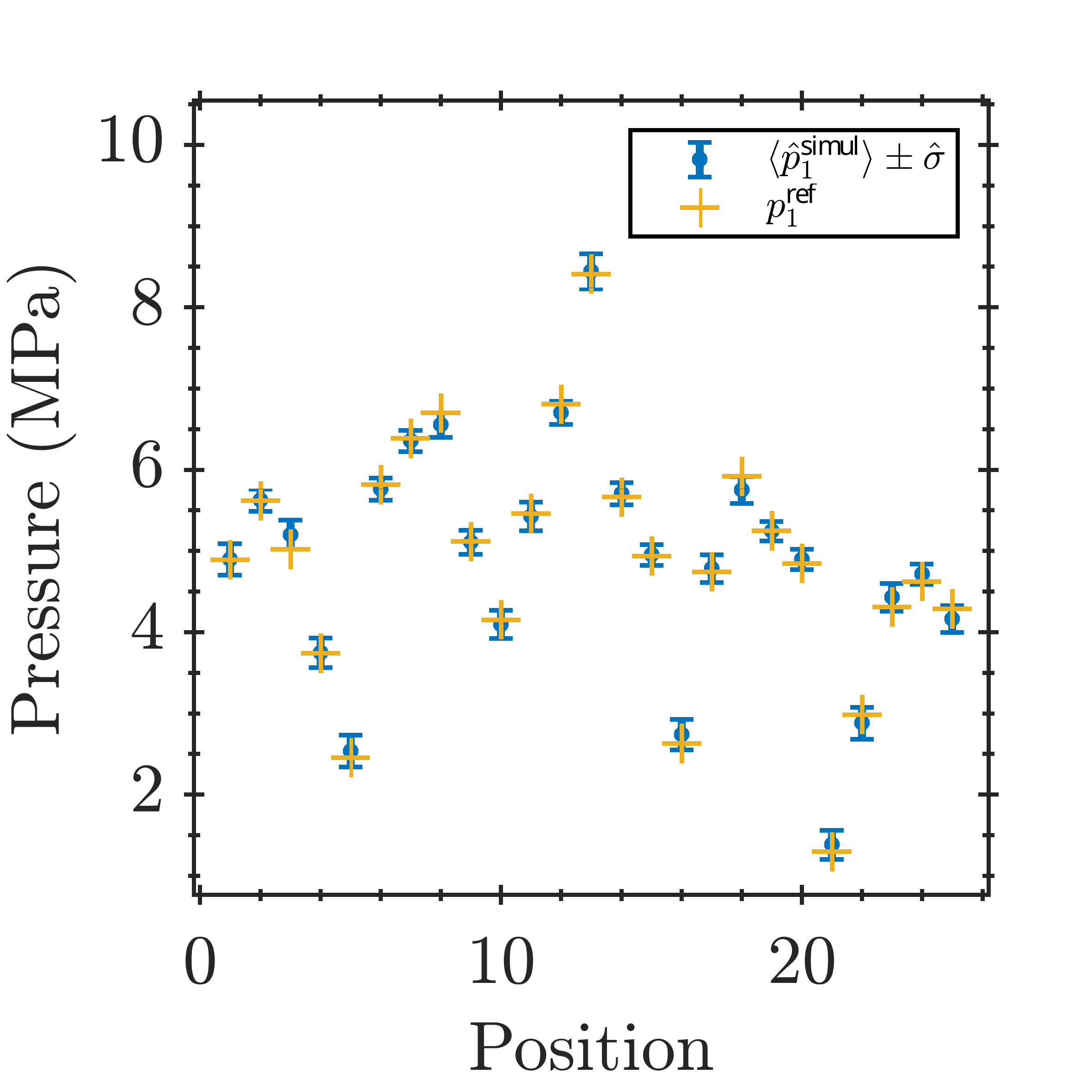}}
    \subfigure[\vaeii]{\includegraphics[width=0.32\linewidth]{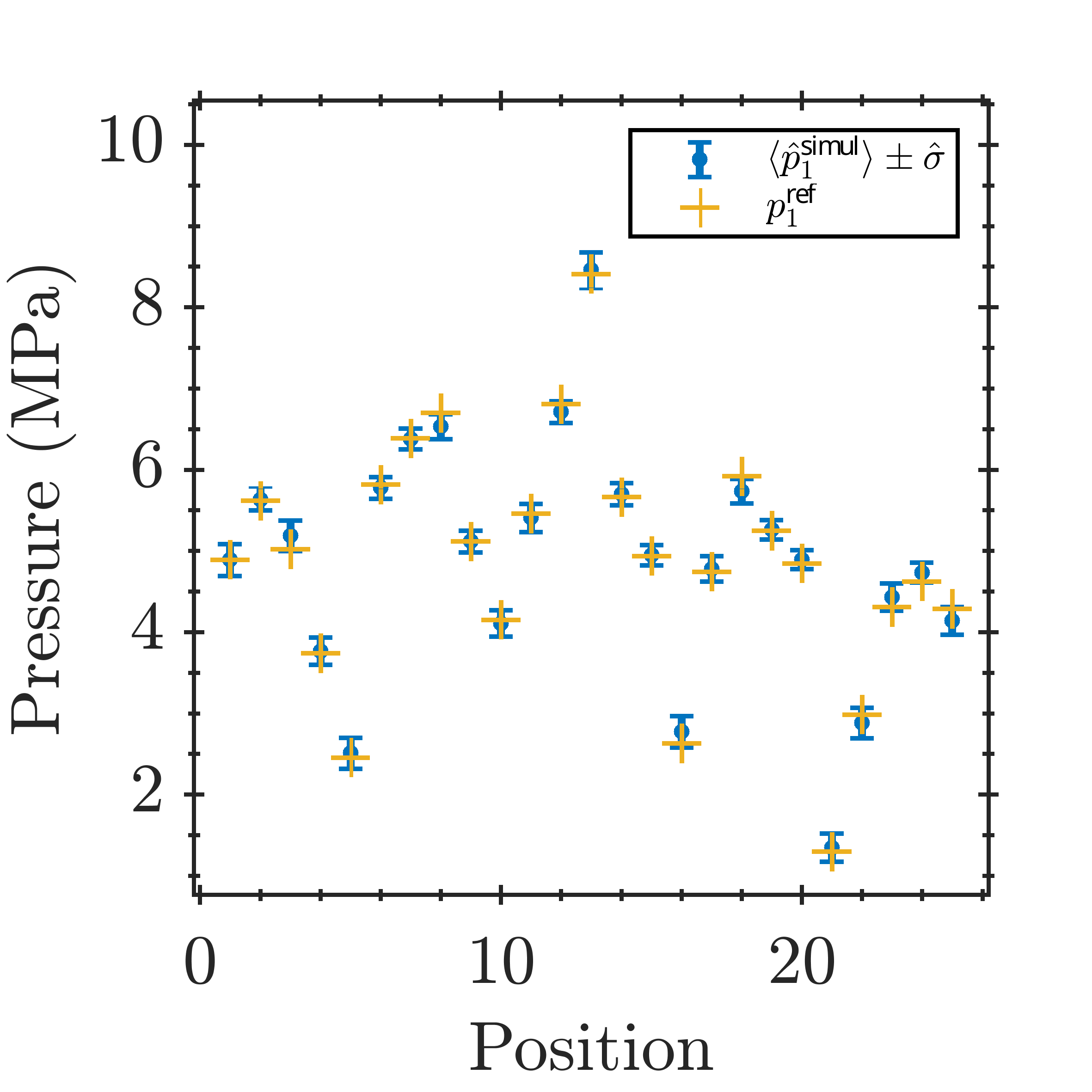}}
    \subfigure[\vaeiii]{\includegraphics[width=0.32\linewidth]{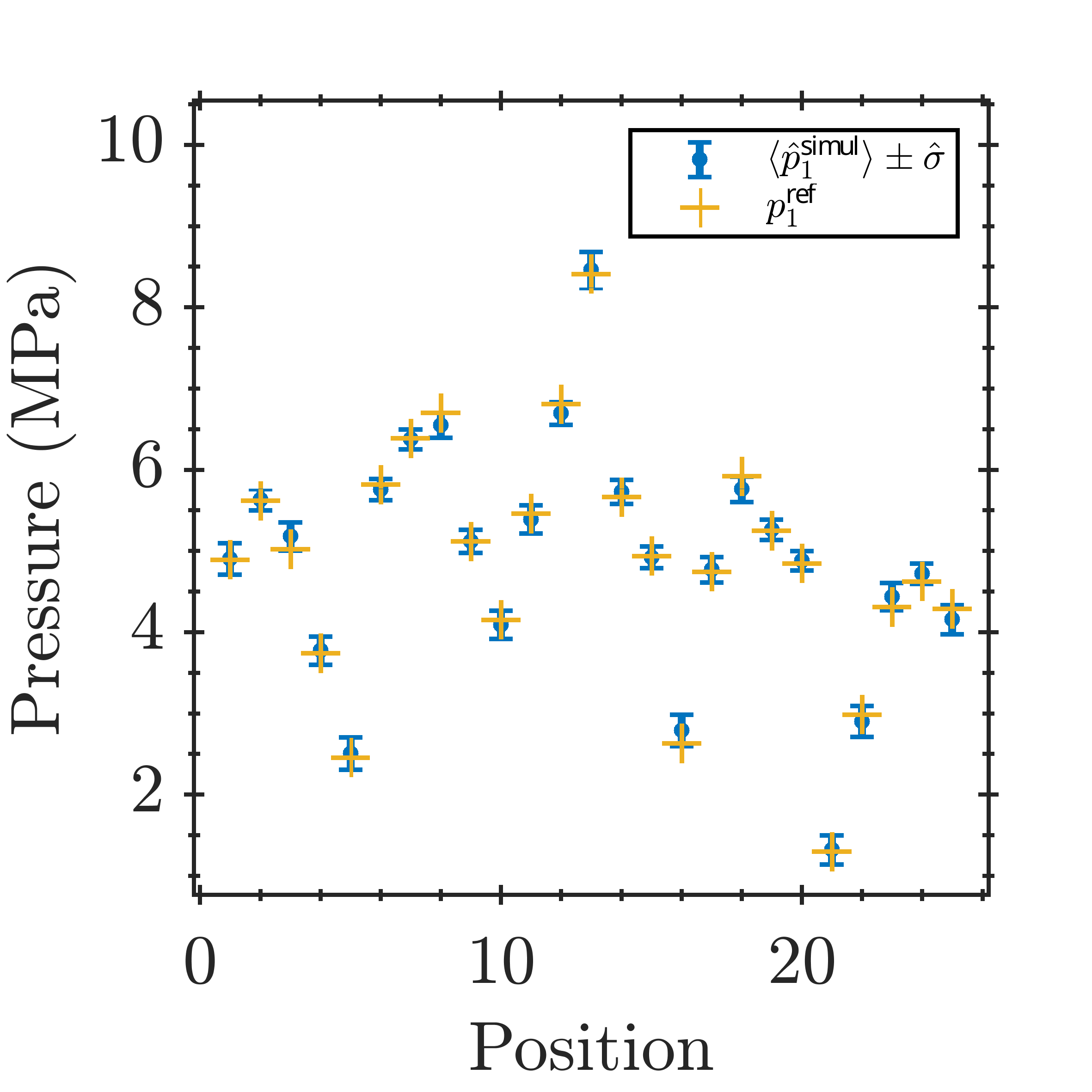}} \\
    \caption{Pressure data along with the recovered pressure. Crosses indicate reference values, circles represent the mean, and bars show the standard deviation.}
    \label{fig:pressure}
\end{figure}

We now begin to explore the posterior distributions obtained in each experiment. \fig{fig:pressure} presents the pressure results in the sensors obtained for the posterior distribution samples. The plots show good agreement between the reference and simulated data, except for some points in the \klen{30} result.

To better compare the experiments, we define the relative error of the data ($\errpostn{j}$) based on the samples of the posterior distributions as

\begin{equation}
    \errpostn{j} = \sqrt{\mynorm{\datan{\rf}\!\left(\vx_{i} \right) - \datapn{\simul}_{j}\!\left(\vx_{i} \right)}} , \qquad \forall \ j=1,2,\dots,\Np,
    \label{eq:errpost}
\end{equation}

\noindent where $\datapn{\simul}_{j}\!\left(\vx_{i} \right)$ is the simulated pressure at sensor in position $\vx_{i}$ for the $j$ sample from posterior distribution. 

The estimates for the mean and standard deviation of the $\errpost$, obtained in each experiment, are presented in the \tab{tab:AR}. We employ the Kolmogorov-Smirnov (\KS) test \citep{Kolmogorov:1933, Smirnov:1948} to compare each experiment's $\errpost$ distributions with those from the \klen{20}, considered the reference experiment. Samples of size $100$ were randomly drawn from the posterior distribution to conduct the test. The $p$-values obtained are displayed in \tab{tab:AR}. A significant difference was observed between the standard experiment (\klen{20}) and both \klen{10} and \klen{30}. This result indicates that selecting an inappropriate correlation length for the prior distribution can lead to a different posterior distribution, even though convergence has been verified. The \VAE\ method experiments did not yield the same results, even in the \vaeii\ case, where the fields with the true correlation length were not in the training set. This demonstrates the proposal's greater adaptability.

\begin{table}[htbp]
\tiny
    \centering
    \caption{Acceptance rate, number of iterations before convergence, and statistical analysis.}
    \label{tab:AR}
    \begin{tabular}{l|c|c|c|c|c}
    \hline\hline
         \textbf{Experiment} & \makecell{\textbf{Estimate of}\\ \textbf{mean acceptance} \\ \textbf{rate ($\eAR$)} ($\%$)}  & \makecell{\textbf{Iterations until}\\ \textbf{convergence}\\ $\left(\psrf<1.2\right)$} & \makecell{\textbf{Estimate}\\[-0mm] \textbf{of the mean}\\ $\left(\hat{\mu}_{_{\errpost}}\right)$} & \makecell{\textbf{Estimate of the}\\[-0mm] \textbf{standard deviation}\\ $\left(\hat{\sigma}_{_{\errpost}}\right)$}
         & \makecell{\textbf{Kolmogorov-}\\[-0mm] \textbf{Smirnov test} \\ ($p$-\textbf{value})}\\
    \hline
        \klen{10} & $16.0$ & $59,\!347$ & $3.8\times 10^{-02}$ & $5.6\times 10^{-03}$ & $0.001$\\
        \klen{20} & $11.8$ & $34,\!136$ & $3.5\times 10^{-02}$ & $5.5\times 10^{-03}$ & $-$\\
        \klen{30} & $10.9$ & $21,\!953$ & $4.2\times 10^{-02}$ & $4.4\times 10^{-03}$ & $<0.001$\\
        \vaei     & $13.1$ & $78,\!740$ & $3.5\times 10^{-02}$ & $5.5\times 10^{-03}$ & $0.260$\\
        \vaeii    & $11.4$ & $76,\!319$ & $3.5\times 10^{-02}$ & $5.4\times 10^{-03}$ & $0.556$\\
        \vaeiii   & $12.7$ & $80,\!879$ & $3.5\times 10^{-02}$ & $5.5\times 10^{-03}$ & $0.140$\\
    \hline\hline
    \end{tabular}
\end{table}

\begin{figure}[htbp]
    \centering
    \subfigure[\klen{10}]{\includegraphics[width=0.32\linewidth]{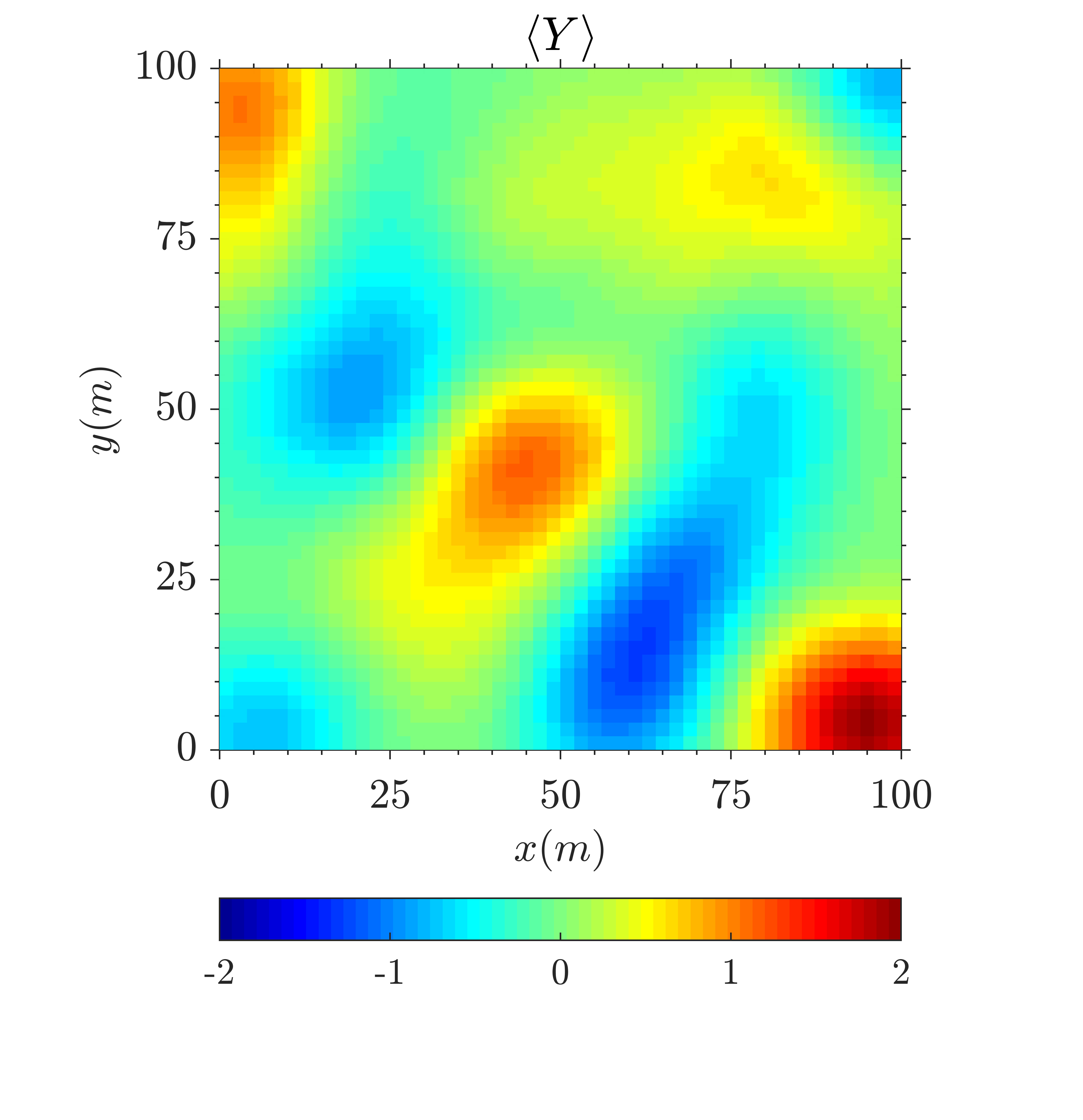}}
    \subfigure[\klen{20}]{\includegraphics[width=0.32\linewidth]{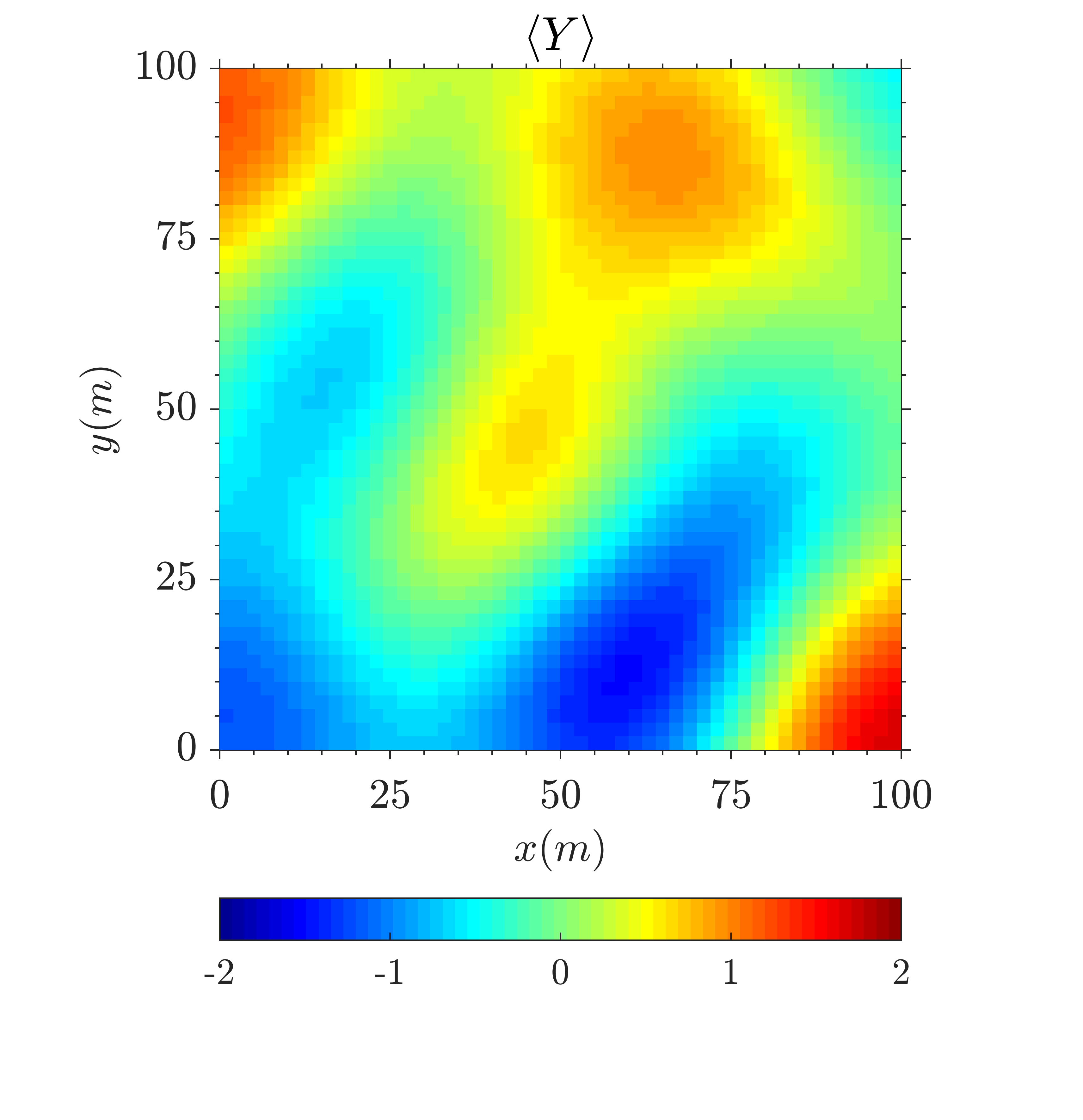}}
    \subfigure[\klen{30}]{\includegraphics[width=0.32\linewidth]{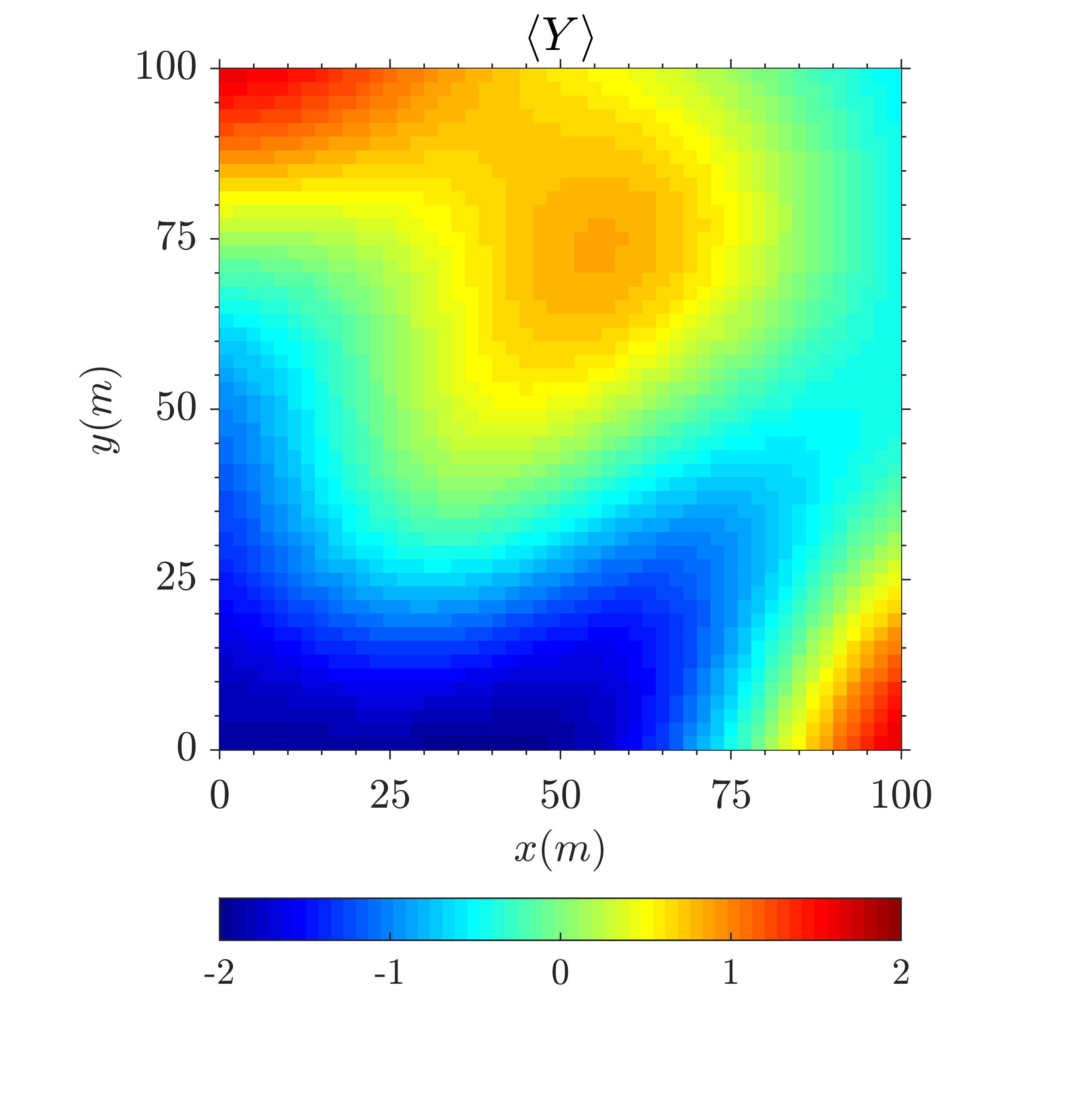}} \\
    \subfigure[\vaei]{\includegraphics[width=0.32\linewidth]{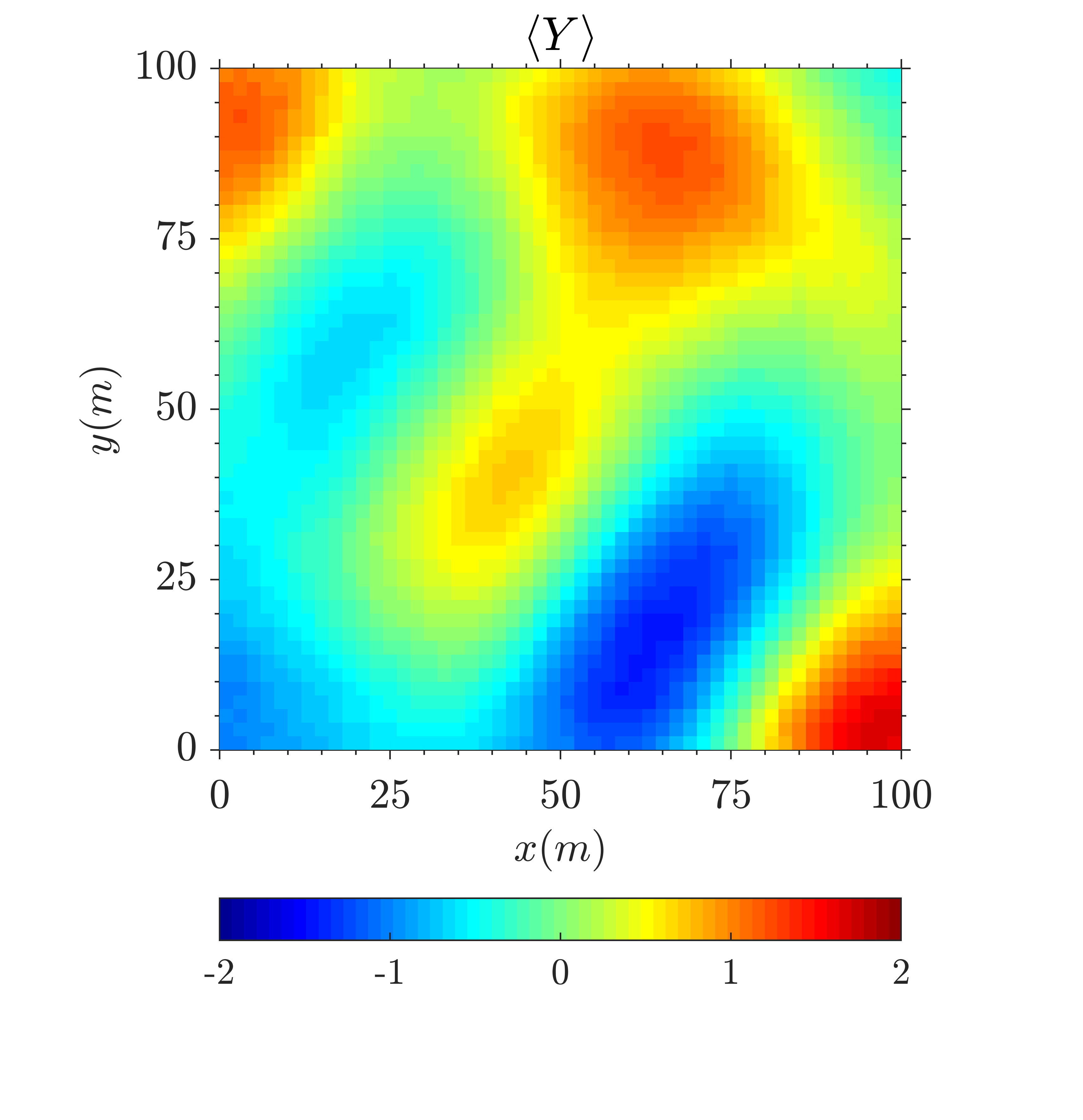}}
    \subfigure[\vaeii]{\includegraphics[width=0.32\linewidth]{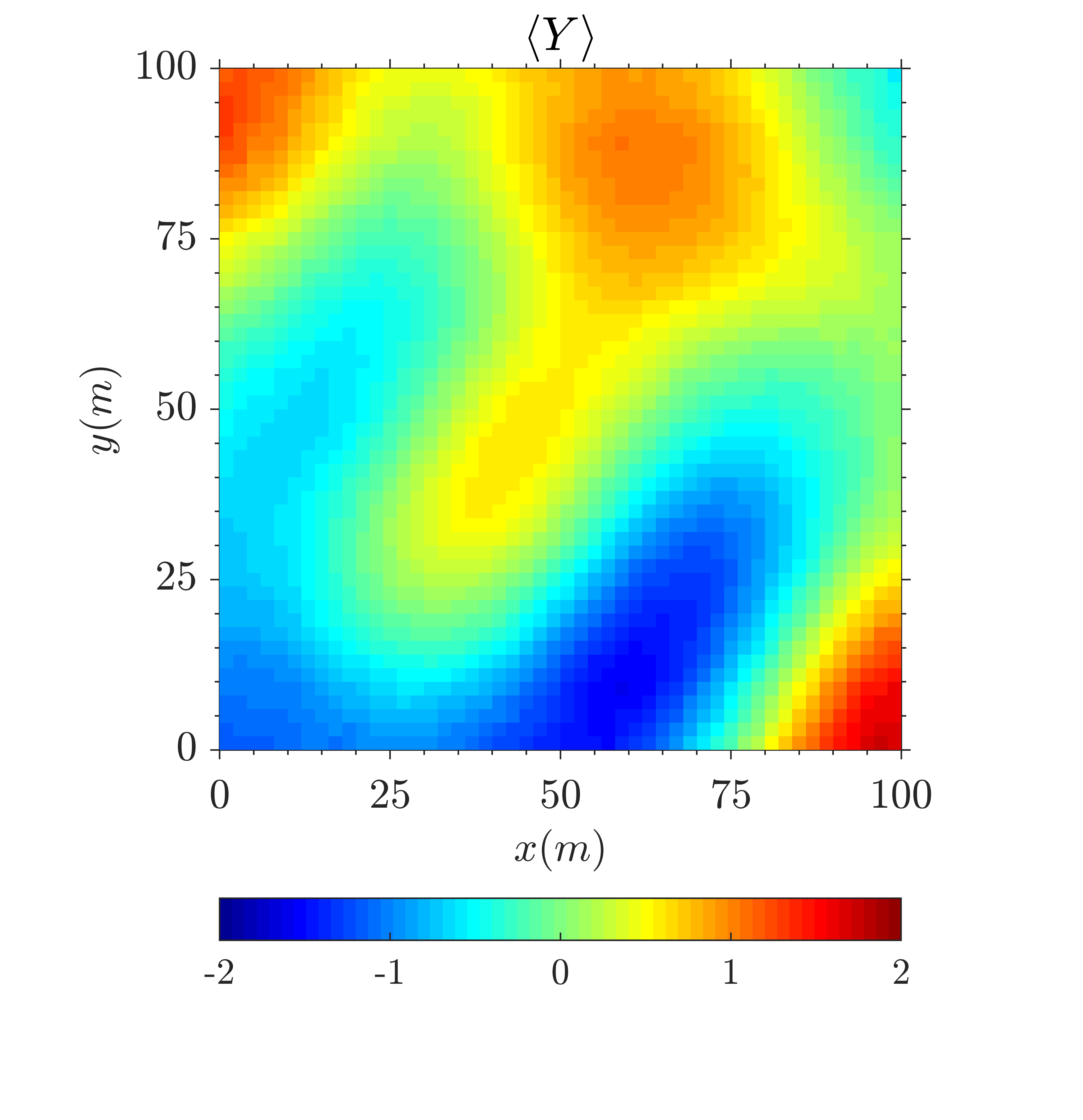}}
    \subfigure[\vaeiii]{\includegraphics[width=0.32\linewidth]{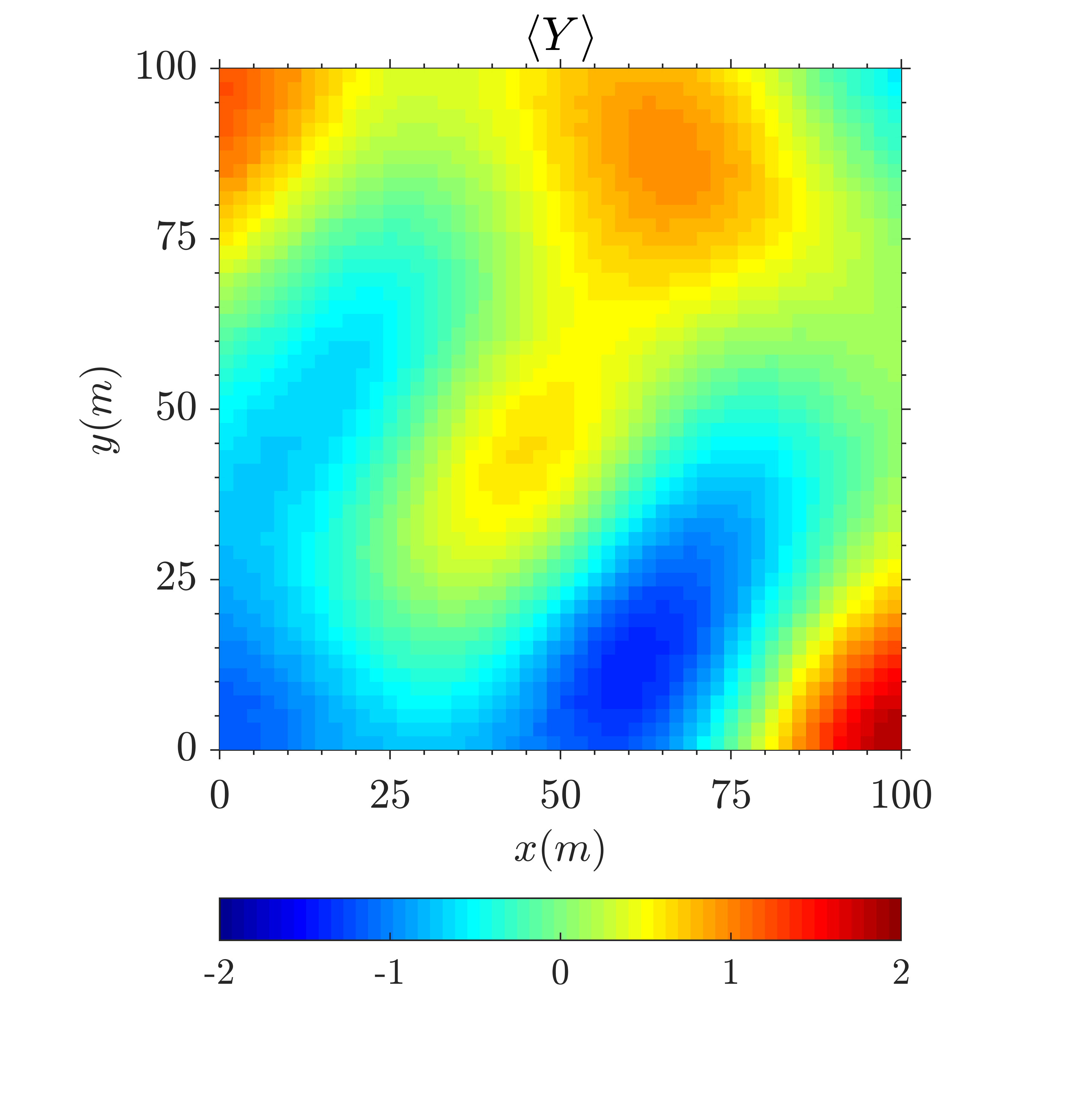}} \\
    \caption{Mean posterior field obtained by each experiment after convergence.}
    \label{fig:meanfield}
\end{figure}

\fig{fig:meanfield} depicts, for each experiment, the estimated mean Gaussian field of the posterior distribution using $5,\!000$ samples. Visually, these fields confirm the last conclusion. The results to \klen{20} and all \VAE\ are very similar.

To enable a quantitative comparison between the reference $\Yref$ (\fig{fig:refsa}) and the posterior distribution fields, we define the relative error as

\begin{equation}
    \errYn{j} = \sqrt{\mynorm{\Yref - \tilde{\Y}_{j}}} , \qquad \forall \ j=1,2,\dots,\Np.
    \label{eq:errY}
\end{equation}

\noindent where $\tilde{\Y}_{j}$ are sampled from the posterior. \tab{tab:ERY} presents the estimated mean and standard deviation of $\errY$. Again, we drew samples (of size $100$) from each posterior distribution to perform the \KS\ test. The results corroborate the last ones. Again, the \KS\ test only identified significant differences between \klen{20} and both \klen{10} and \klen{30}. 

\begin{table}[htbp]
\tiny
    \centering
    \caption{Relative error between the reference and posterior Gaussian fields.}
    \label{tab:ERY}
    \begin{tabular}{l|c|c|c}
    \hline\hline
         \textbf{Experiment} & \makecell{\textbf{Estimate of the}\\[-0mm] \textbf{mean} $\left(\hat{\mu}_{_{\errY}}\right)$} & \makecell{\textbf{Estimate of the standard}\\[-0mm] \textbf{deviation} $\left(\hat{\sigma}_{_{\errY}}\right)$}
         & \makecell{\textbf{Kolmogorov-}\\[-0mm] \textbf{Smirnov} ($p$-\textbf{value})}\\
    \hline
        \klen{10} & $1.1\times 10^{+00}$ & $1.4\times 10^{-01}$ & $<0.001$\\
        \klen{20} & $8.7\times 10^{-01}$ & $1.5\times 10^{-01}$ & $-$\\
        \klen{30} & $9.9\times 10^{-01}$ & $1.4\times 10^{-01}$ & $<0.001$\\
        \vaei     & $8.4\times 10^{-01}$ & $1.4\times 10^{-01}$ & $0.556$\\
        \vaeii    & $8.4\times 10^{-01}$ & $1.3\times 10^{-01}$ & $0.443$\\
        \vaeiii   & $8.6\times 10^{-01}$ & $1.3\times 10^{-01}$ & $0.344$\\
    \hline\hline
    \end{tabular}
\end{table}

\section{Conclusions} \label{sec:conclusions}

The Variational Autoencoder (\VAE) method can learn a wide range of field properties from training data, thus enhancing its applicability to real-world scenarios. This flexibility allows for a more versatile prior distribution in Markov Chain Monte Carlo (\mcmc) methods, as it eliminates the strict requirement of knowing the covariance function in advance. This is particularly advantageous in Bayesian inverse problems, where such information is often unavailable.

This work demonstrates that the \VAE\ approach yields results comparable to the Karhunen-Loève Expansion (\kle) when the known correlation length is applied. However, it significantly outperforms \kle\ when the assumed correlation length deviates from the actual value, making it a more robust choice in practical situations. Additionally, the \VAE\ method effectively reduces the stochastic dimensionality of the problem, which leads to faster convergence of \mcmc\ simulations while maintaining accuracy.

The \mcmc\ simulations that use \VAE-based priors showed favorable convergence behavior and improved efficiency, confirming the validity of the method for Bayesian inference in high-dimensional inverse problems. This study underscores the advantages of employing deep generative models like \VAE\ to enhance \mcmc\ methods, providing a more flexible and computationally efficient approach to solving Bayesian inverse problems in porous media and other complex domains.

\appendix


\section*{Acknowledgment}
The authors acknowledge the National Laboratory for Scientific Computing (LNCC\-/\-MCTI, Brazil) for providing HPC resources from the \textit{SDumont} supercomputer and contributing to this article's research results. URL: \url{http://sdumont.lncc.br}. They are also grateful to Fabio Porto from the Data Extreme Lab (DEXL) research group for the GPU resources used to train the autoencoders. URL: \url{https://dexl.lncc.br}. This study was financed in part by the {\em Coordena\c{c}\~ao de Aperfei\c{c}oamento de Pessoal de N\'{\i}vel Superior} - Brazil (CAPES) – Finance Code 88887.932390/2024-00. This material is based upon work supported by the National Science Foundation under Grant No. 2401945. Any opinions, findings, and conclusions or recommendations expressed in this material are those of the authors and do not necessarily reflect the views of the National Science Foundation. The work of F.P. is also partially supported by The University of Texas at Dallas Office of Research and Innovation through the SPARK program.


\clearpage

\end{document}